\documentclass[10pt,journal,compsoc]{IEEEtran}

\ifCLASSOPTIONcompsoc
  \usepackage[nocompress]{cite}
\else
  \usepackage{cite}
\fi

\usepackage[pdftex]{graphicx}
\usepackage{subfig}
\usepackage{booktabs}
\usepackage{tabularx} 
\usepackage{amsmath}
\begin{document}
\title{Context-Aware Mixed Reality: A Framework for Ubiquitous Interaction}
\author{Long~Chen,
        Wen~Tang,
       Nigel~John,
        Tao Ruan Wan,
        Jian Jun Zhang 
\IEEEcompsocitemizethanks{\IEEEcompsocthanksitem L. Chen and W. Tang are with the Department
of Creative Technology, Bournemouth University, Poole, UK, BH12 5BB.}
\IEEEcompsocitemizethanks{\IEEEcompsocthanksitem N. John is with the Department of Computer Science, University of Chester, Chester, UK, CH1 4BJ.}
\IEEEcompsocitemizethanks{\IEEEcompsocthanksitem T. R. Wan is with the School of Informatics at the University of Bradford, UK, BD7 1DP}
\IEEEcompsocitemizethanks{\IEEEcompsocthanksitem J. J. Zhang is with the National Centre for Computer Animation at the Bournemouth University, Poole, UK, BH12 5BB.}
}


\IEEEtitleabstractindextext{%
\begin{abstract}
Mixed Reality (MR) is a powerful interactive technology that yields new types of user experience. We present a semantic based interactive MR framework that exceeds the current geometry level approaches, a step change in generating high-level context-aware interactions. Our key insight is to build semantic understanding in MR that not only can greatly enhance user experience through object-specific behaviours, but also pave the way for solving complex interaction design challenges. The framework generates semantic properties of the real world environment through dense scene reconstruction and deep image understanding. We demonstrate our approach with a material-aware prototype system for generating context-aware physical interactions between the real and the virtual objects. Quantitative and qualitative evaluations are carried out and the results show that the framework delivers accurate and fast semantic information in interactive MR environment, providing effective semantic level interactions. 
\end{abstract}

}

\maketitle

\IEEEdisplaynontitleabstractindextext

\IEEEpeerreviewmaketitle

\IEEEraisesectionheading{\section{Introduction}\label{sec:introduction}}

\IEEEPARstart{M}{ixed} Reality combines computer vision with information science, and computer graphics as a cross-cutting technology. It makes seamless connections between virtual space and the real world, by not only superimposing computer-generated information onto the real world environment, but also making progress on novel user interaction for new experience. This interactive technology will soon become ubiquitous in many applications, ranging from personal information systems, industrial and military simulations, office use, digital games to education and training. 

\begin{figure}[!t]
\centering
\subfloat[]{\includegraphics[width=0.24\textwidth]{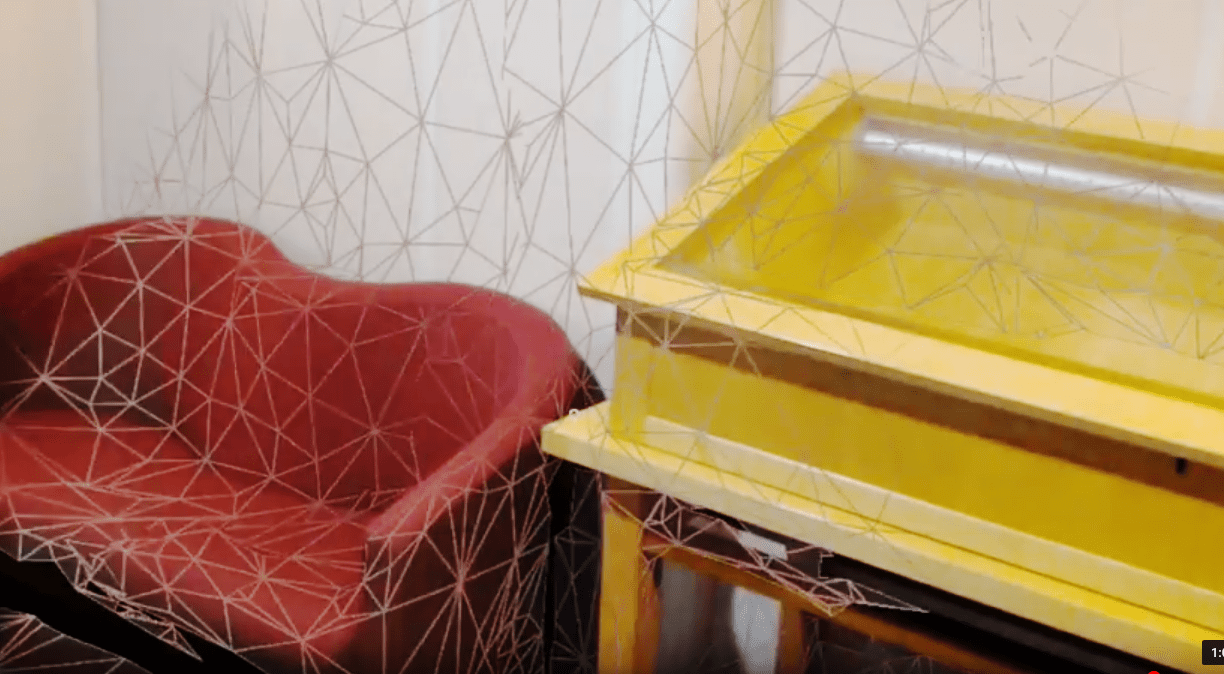}}\
\subfloat[]{\includegraphics[width=0.24\textwidth]{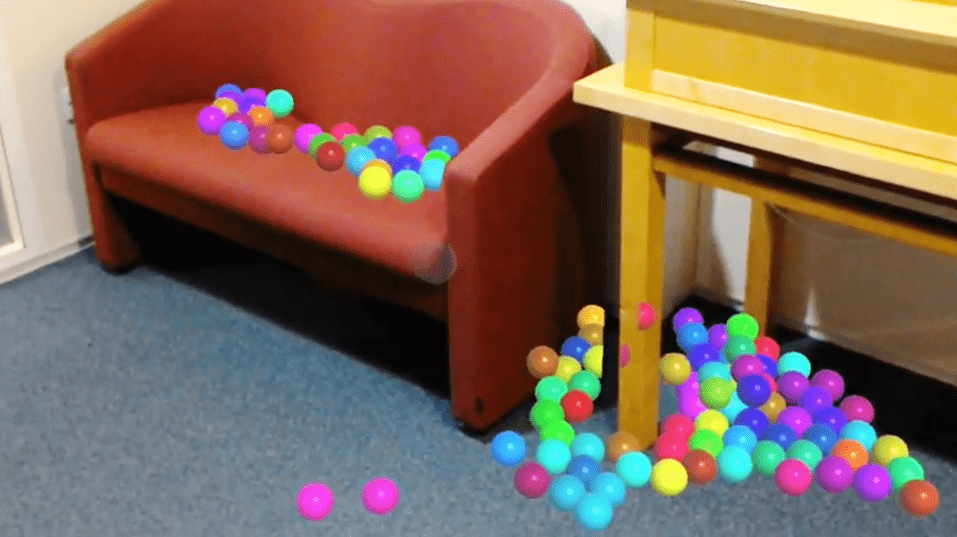}}\\
\subfloat[]{\includegraphics[width=0.24\textwidth]{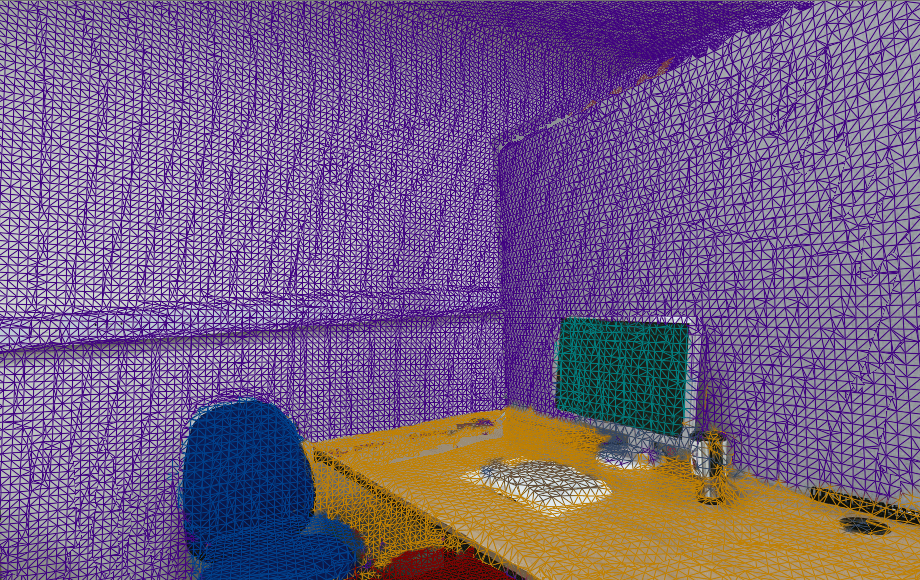}}\
\subfloat[]{\includegraphics[width=0.24\textwidth]{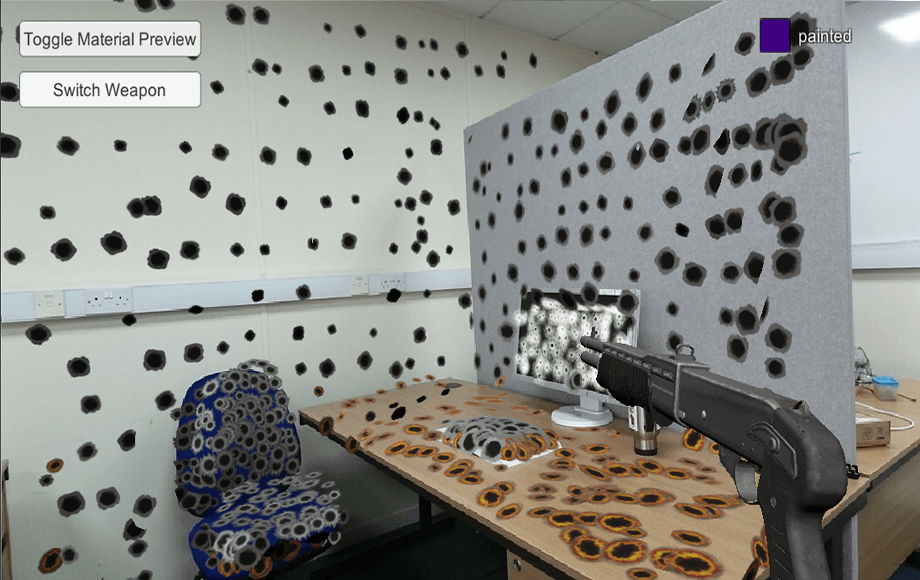}}\\
\caption{(a): Microsoft HoloLens is capable of reconstructing the environment by its built-in ''spatial mapping'' function and provide a geometric mesh for geometry based interaction.  (b): The Ball Pit MR games based on geometry interaction for Microsoft HoloLens. (c) Our proposed framework can provide semantic mesh for more advanced context-aware interaction. (d) A shooting game developed based on our proposed framework. Note that the bullet holes are different according to the objects' properties.}
\label{intro}       
\end{figure}

The latest research on Simultaneous Localisation and Mapping (SLAM) has opened up a new world for MR development, greatly increased the camera tracking accuracy and robustness. The sparse SLAM systems~\cite{Davison2007}~\cite{Klein2007}~\cite{Mur-ArtalMontielTardos2015} are proven to be efficient 3D tracking methods for monocular cameras, but structural information are absent from these systems. In contrast, dense SLAMs~\cite{Newcombe2011a}~\cite{Newcombe2011}~\cite{Newcombe2015} construct dense surfaces to generate geometric information of the real scene, enabling geometric interactions in MR environment. The collision effects between virtual and real-world objects in these geometry-aware MR systems do increase the immersion of the user experience (as can be seen in Figure \ref{intro} (a) and (b) for the Ball Pit MR game in Microsoft HoloLens). However, as individual semantic properties of various different objects of the real world remain undetected, geometry-aware MRs are unable to distinguish different object properties and may always generate uniform interactions between each other, which will break the continuous user experience in MR~\cite{Grubert2017}.

The natural first step moving away from purely geometric-based approaches towards generating context-aware interactions is to understand the real environment semantically in MR. Semantic segmentation \cite{Garcia-Garcia2017} \cite{Shelhamer2017} \cite{crfasrnn_iccv2015} \cite{Chen2017} \cite{Badrinarayanan2017} leading to semantic understanding is not new to computer vision. However, very few prior works of utilizing semantic information in MR are reported. Semantic understanding in MR presents additional challenges (1) associating semantics with the structural information of the environment seamlessly on-the-go and (2) retrieving the semantics then generating appropriate interactions.


Embedding semantic information extracted from a 2D image space into the 3D structure of a MR environment is hard, because of the required high accuracy. Careful considerations are needed when designing semantic-based MR interactions. 

Realistic interactions in MR require not only geometric and structural information, but also semantic understandings of the scene. While geometric structure allows accurate information augmentation and placement, at the user experience level, semantic knowledge will enable realistic interactions between the virtual and real objects. For example realistic physical interactions (e.g. a virtual glass can be shunted on a real concrete floor) in MR. More importantly, using semantic scene descriptions, we can develop high-level tools for efficient design and constructions of large and complex MR applications.

With the gap that this work addresses, we propose a novel context-aware semantic MR framework and demonstrate its effectiveness through example interactive applications. We show that how an end-to-end Deep Learning (DL) framework and the dense Simultaneous Localisation and Mapping (SLAM) can be used for semantic information integration in MR environment and how context-aware interactions can be generated. We present the labelling of material properties of the real environment in 3D space as a novel example application to deliver realistic physical interactions between the virtual and real objects in MR. To the best of our knowledge, this is the first work to present a context-aware MR (1) using deep learning based semantic scene understanding and (2) generating semantic interactions at the object-specific level, one step further towards the high-level conceptual interaction modelling in complex MR environment for enhanced user experience. 

Dense SLAM KinectFusion \cite{Newcombe2011} is used for camera pose recovery and 3D model reconstruction for creating a classic geometry-aware MR environment first. We trained a Conditional Random Fields as Recurrent Neural Networks (CRF-RNN) \cite{crfasrnn_iccv2015} using a large-scale material database \cite{minc} for detecting material properties of each object in the scene. The 3D geometry/model of the scene is then labelled with the semantic information about the materials made up of the scene, so that realistic physics can be applied during interactions via real-time inference to generate corresponding physical interactions based on the material property of the object that the user is interacting with, as shown in a shooting game example for object-specific material-aware interactions, Figures \ref{intro} (c) and (d). 

The framework is both efficient and accurate in semantic labelling and inference for generating realistic context-aware interactions. Two tests are designed to evaluate the effectiveness of the framework (1) an accuracy study with an end-to-end system accuracy evaluation by comparing the dense semantic ray-casting results with manually labelled ground truth from 25 keyframes of two different scenes and (2) a user experiment with 68 participants to qualitatively evaluate user experience using three different MR conditions. The results show that the framework delivers more accuracy in 3D semantic mappings than directly using the state-of-the-art 2D semantic segmentation. The proposed semantic based interactive MR system ($M=8.427, SD=1.995$) has a significant improvement ($p<0.001$) on the realism and user experience than the existing MR system approaches that do not encode semantic  descriptions and context-aware interaction modelling ($M=5.935, SD=1.458$). 

In the next section, we review related work on geometry-based MR Interactions, and recent approaches to semantic segmentations using Convolutional Neural Network. The following sections introduce our framework with SLAM dense reconstructions of the scene and the 3D semantic fusion, and describe our implementation and evaluation framework. Finally, we demonstrate our results compared with the state-of-the-art semantic segmentation algorithms.


\begin{figure*}[bt]
\centering
\includegraphics[width=0.95\textwidth]{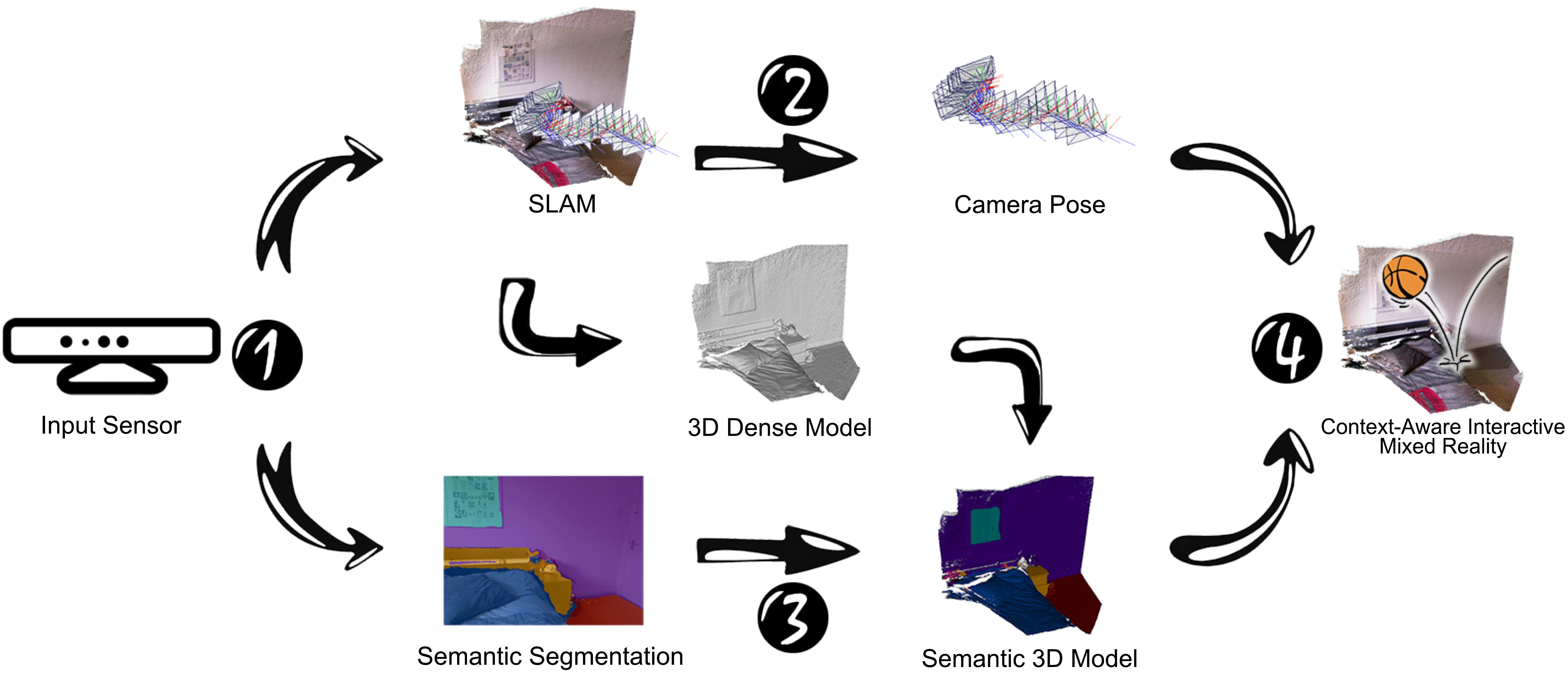}
\caption{Flowchart demonstrates the whole framework.}
\label{flowchart}       
\end{figure*}

\section{Previous Work}

Our approach draws on recent success of dense SLAM algorithms~\cite{Newcombe2011a}~\cite{Newcombe2011}~\cite{Newcombe2015} and deep learning for semantic segmentations \cite{Garcia-Garcia2017} \cite{Shelhamer2017} \cite{crfasrnn_iccv2015} \cite{Chen2017} \cite{Badrinarayanan2017} that have been mostly used in the field of robotics until now.

\subsection{Geometry-based MR Interaction}

Interaction modelling between virtual and real objects in MR are mostly geometry-based through plane feature detections or full 3D reconstructions of the real-world. Methods of using plane detections~\cite{Salas-Moreno2014}~\cite{SnapToReality2016} estimate planar surfaces in the real-world, onto which virtual objects are placed and collided with. Random Sample Consensus (RANSAC) algorithm~\cite{Fischler1981} estimates planar surfaces based on sparse 3D feature points extracted from a monocular camera. Plan detections require no depth cameras, are computationally efficient and run on mobile phones. Mobile MR experience is shown in the newly released Mobile AR systems~\cite{Apple2017}~\cite{Google2017}. One obvious shortcoming of plane detections is the requirement for large planar surfaces to delivery MR interactions. Collision meshes for non-planar surfaces are impossible, hence,  restricting user experience to the area and types of objects which users can interact with. 

Recent advances in depth sensors, display technologies and SLAM software~\cite{Newcombe2011a}~\cite{Newcombe2011}~\cite{Whelan2013}~\cite{Newcombe2015} have opened up the potential of MR systems. Spatial structures of the real environment can be generated at ease to provide accurate geometries for detecting collisions between virtual and real objects. We saw examples of geometry-based interactions e.g. a virtual car 'drives' on an uneven real desk~\cite{Newcombe2011a}; the Super Mario game played on real building blocks~\cite{Kim2013}; and the Ball Pit game in HoloLens~\cite{Microsoft2017}. Figures 1 (a) and (b) illustrate the concept. Impressive as they are, the state of the art systems are still limited to the basic and uniform geometry-based virtual and real object interactions. Without high-level semantic descriptions and scene understandings, continuous user experience in MR is compromised and easily broken, and the realism and immersion are reduced. One example is in the Ball Pit game, material properties of the real objects are not recognized, thus a ball falling onto a soft surface would still bounce back unrealistically against the law of physics.  

\subsection{Deep Semantic Understanding}

Semantic segmentation is an emerging technology in computer vision. The recent success of Convolutional Neural Network(CNN) has achieved the semantic level image recognition and classification with great accuracy~\cite{NIPS2012_4824}, enabling many novel applications. In last few years more complex neural networks such as FCN \cite{Shelhamer2017}, CRF-RNN \cite{crfasrnn_iccv2015}, DeepLab \cite{Chen2017} and SegNet \cite{Badrinarayanan2017} have enabled image understanding at the pixel level. Semantic information at every pixel of an image can be predicted and labelled when using these neural networks trained on large-scale databases. 

Combined with SLAM systems, semantic segmentation can be achieved in 3D environments~\cite{cofusion} \cite{Tateno2017CNNSLAMRD} \cite{Zhao2017} \cite{McCormac2016}, a promising future in robotic vision understanding and autonomous driving. Unlike these existing methods that aimed at providing semantic understanding of the scene for robots, we focus on human interactions. Our goal is to provide users with realistic semantic level interactions in MR. In this paper, for the first time, we use MR as a bridge to connect AI and human for a better understanding of the world via intelligent context-aware interaction. 

\subsection{Context and Semantic awareness in XR environment}

Prior approaches have studied context and semantic understandings in 3D virtual environment, e.g. semantic inferring in interactive visual data exploration~\cite{North2012}; enhancing software quality for multi-modal Virtual Reality (VR) systems~\cite{Fischbach2017}; visual text analytics~\cite{Endert2012}; and interactive urban visualization \cite{Deng2016}. Context awareness is also introduced in computer-aided graphic design such as inbetweening of animation\cite{Yang2018}; 3D particle clouds selection\cite{Yu2016}; and illustrative volume rendering \cite{Rautek2007}. Virtual object classifications are proposed in VR applications by using semantic associations to describe virtual object behaviours \cite{Chevaillier2012}. The notion of \textit{conceptual modelling} for VR applications is pointed out by Troyer \textit{et al.}, highlighting a large gap between the conceptual modelling and VR implementations. It is suggested to take a phased approach (i.e. conceptual specification, mapping and generation phases) to bridge the gap~\cite{DeTroyer2007}. 

Recently, the idea of extending Augmented Reality (AR) applications to become context-aware has appeared in computer graphics~\cite{Grubert2017}, which proposes to classify context sources and context targets for continuous user experience. A method is proposed for authentically simulating outdoor shadows to achieve seamless context-aware integration between virtual and real objects for mobile AR~\cite{Barreira2018}. 

We address ubiquitous interactions in MR environment and see deep semantic understanding of the environment as the first step towards the high-level interaction design for MR. Real-time 3D semantic reconstruction is an active research topic in robotics with many recent works being focused on object semantic labelling. We now re-design and fine-tune the architecture for this MR framework.

\section{Framework Overview}

Figure \ref{flowchart} shows the proposed framework. Starting from an \textbf{\textit{\textcircled{1}Input Sensor}}, two main computation streams are constructed: \textbf{\textit{\textcircled{2}Tracking \& Reconstruction Stream}} and \textbf{\textit{\textcircled{3}Context Detection \& Fusion Stream}}, which are finally merged and output to the  \textbf{\textit{\textcircled{4}Interactive MR Interface}} for generating context-aware virtual-real interactions.

\subsection*{Input Sensor}
An \textit{input sensor}, a RGB-D camera such as Microsoft Kinect, ASUS Xtion series or built-in sensors on Microsoft HoloLens, is used to acquire the depth information directly for the 3D reconstruction of the environment. Monocular or stereo cameras would also work if combined with dense SLAM systems~\cite{Newcombe2011a}, but the accuracy and real-time performance of Mono devices are not guaranteed.

\subsection*{Camera Tracking \& Reconstruction Stream}
The \textit{tracking \& reconstruction stream} shown in the upper path of Figure \ref{flowchart} processes the video captured by the input sensor. A SLAM system continuously estimates the camera pose while simultaneously reconstruct a 3D dense model. This is a typical method used in latest MR systems such as Microsoft HoloLens for implementing geometry-aware MR. A dense 3D model is served as the spatial collision mesh and the inverse of the camera pose extracted from SLAM guides the movement of the collision mesh to visually correspond to the real-world objects.

\subsection*{Context Detection \& Fusion Stream}
The lower path of Figure \ref{flowchart} shows the \textit{Context Detection Stream}. 2D image sequences from the input sensor are context sources to be processed by semantic segmentation algorithms that can densely output the pixel-wise object attributes and properties of the scene. Based on the semantic segmentation information, the context information relevant for implementing context-aware experience is generated. Then the 2D semantic segmentation results are projected onto the scene and fused with the 3D dense model (from \textit{tracking \& reconstruction stream}) to generate a semantic 3D model based on the camera pose of the current frame.

\subsection*{Interactive MR Interface}
The semantic 3D model are combined with the camera pose to provide a context-aware MR environment. High-level interactions can be designed based on the semantics. Furthermore, tools can be developed to facilitate design and automatic constructions of complex MR interactions in different applications.

The advantages of the proposed framework are: 

    1) \textbf{Accurate 3D Semantic Labeling}: The Context Detection \& Fusion Stream can predict a pixel-wise segmentation of the current frame, which is further fused onto the 3D dense model. The semantic 3D model is generated with each voxel contains the knowledge of the context information of the environment. The voxel-based context-aware model delivers the semantic information through ray-cast queries about the object properties in order to generate different interactions.  
    Object properties can be high-level descriptions, for example types of material and interaction attributes.
    
    2) \textbf{Real-time Performance}: In deep learning based approaches the semantic segmentation is computationally expensive especially for processing frame by frame in real-time applications. We achieve the real-time performance by storing the semantic information into the 3D dense model after the initial segmentation process, so that the semantic segmentation is not processed at every frame, but at certain frames.
    
    3) \textbf{Automatic Interaction Design}: With the context information available, virtual and real object interactions can be designed and computed by feeding the object attributes of the real world to the target software module for processing e.g. a physics module or an agent AI module. For example, realistic physical interactions between virtual and real objects can be computed by feeding the material properties of the real world to physics simulation algorithm (such as our throwing plates game in the following section).


\section{implementation}

We present our novel MR framework in the context of object material-aware interactions as an implementation example to demonstrate the concept of context-aware MR. The material properties in the MR generates realistic physical interactions based on the objects' material property. This example implementation is also used for accuracy study and user experiment.

\subsection{Camera Tracking and Model Reconstruction}


The accurate camera tracking and dense 3D model reconstructions of the environment are achieved by adopting a dense SLAM system~\cite{Newcombe2011}, which estimates camera poses and reconstructs the 3D model in real-time. Depth images from a Kinect sensor are projected into the 3D model. The camera pose and a single global surface model can be simultaneously obtained through a coarse-to-fine iterative closest point (ICP) algorithm. The tracking and reconstruction processes consist four steps: 

    1) Each pixel acquired by the depth camera is firstly transformed into the 3D space by the camera's intrinsic parameters and the corresponding depth value is acquired by the camera; 
    
    2) An ICP alignment algorithm is performed to estimate the camera poses between the current frame and the global reconstructed model;
    
    3) With the available camera poses, each consecutive depth frame can be fused incrementally into one single 3D reconstruction by a volumetric truncated signed distance function (TSDF);
    
    4) Finally, a surface model is predicted via a ray-casting process.

A Microsoft Kinect V2 is used as the input sensor with OpenNI2 driver to capture RGB images and calibrated depth images at the resolution of 960x540 at 30 frames per second.

\subsection{Deep Learning for Material Recognition}
We trained a deep neural network for the 2D material recognition task. Our neural network is implemented in \textit{caffe}~\cite{Jia2014} based on the CRF-RNN  architecture~\cite{crfasrnn_iccv2015}, which combines the FCN with Conditional Random Fields (CRF) based on the probabilistic graphical modelling for contextual boundary refinement. We use the Materials in Context Database (MINC) \cite{minc} as the training database that contains 3 million labelled point samples and 7061 labelled material segmentations in 23 different material categories.

The VGG-16 pre-trained model for ImageNet Large-Scale Visual Recognition Challenge (ILSVRC)~\cite{Simonyan15} is used as the initial weights of our neural network. Based on the MINC dataset, we then fine-tune the network from 1000 different classes (ImageNet contains 1000 classes of labels) to 23 class labels as the output. VGG-16 is a CNN model specifically designed for classification tasks and only produces a single classification result for a single image. Therefore, we manually cast the CNN into a Fully Convolutional Network (FCN) for pixel-wise dense outputs~\cite{Shelhamer2017}. By transforming the last three inner product layers into convolutional layers, the network can learn to make dense predictions efficiently at the pixel level for tasks like semantic segmentation. The fully-connected CRF model is then integrated into FCN to improve the semantic labelling results. 

Fully-connected CRF encodes pixel labels as random variables form a Markov Random Field (MRF)~\cite{Kindermann1980} conditioned on a global observation (the original image). By minimising the CRF energy function in the Gibbs distribution~\cite{Ladicky2009}, we obtain the most probable label assignment for each pixel in an image. With this process, the CRF refines the predicted label using the contextual information. It is also able to refine weak semantic label predictions to produce sharp boundaries and better segmentation results (see Figure \ref{compare} for the comparison of FCN and CRF-RNN). During the training process, CRF is implemented by multiple iterations, each takes parameters estimated from the previous iteration, which can be treated as a Recurrent Neural Network (RNN) structure \cite{crfasrnn_iccv2015}. 

As the error of CRF-RNN can be passed through the whole network during a backward propagation, the FCN can generate better estimations for CRF-RNN optimization process during the forward propagation. Meanwhile, CRF parameters, such as weights of the label compatibility function and Gaussian kernels can be learned from the end-to-end training process.

We use 80\% of the 7061 densely labelled material segmentations in the MINC dataset as the training dataset and the rest of 20\% as testing sets. The training dataset is trained using a single Nvidia Titan X GPU for 50 epochs, after which there is no significant decrease of loss. For testing results, we obtain a mean accuracy of 78.3\% for the trained neural network. The trained network runs at around 5 frames per second for the 2D dense semantic segmentation at the resolution of 480x270 on a Nvidia Titan X GPU. We input 1 frame into our neural network for every 12 frames according to our test to achieve a trade-off between the speed and accuracy.



\subsection{Bayesian Fusion for 3D Semantic Label Fusion}

\begin{figure*}[!t]
\centering
\includegraphics[width=0.98\textwidth]{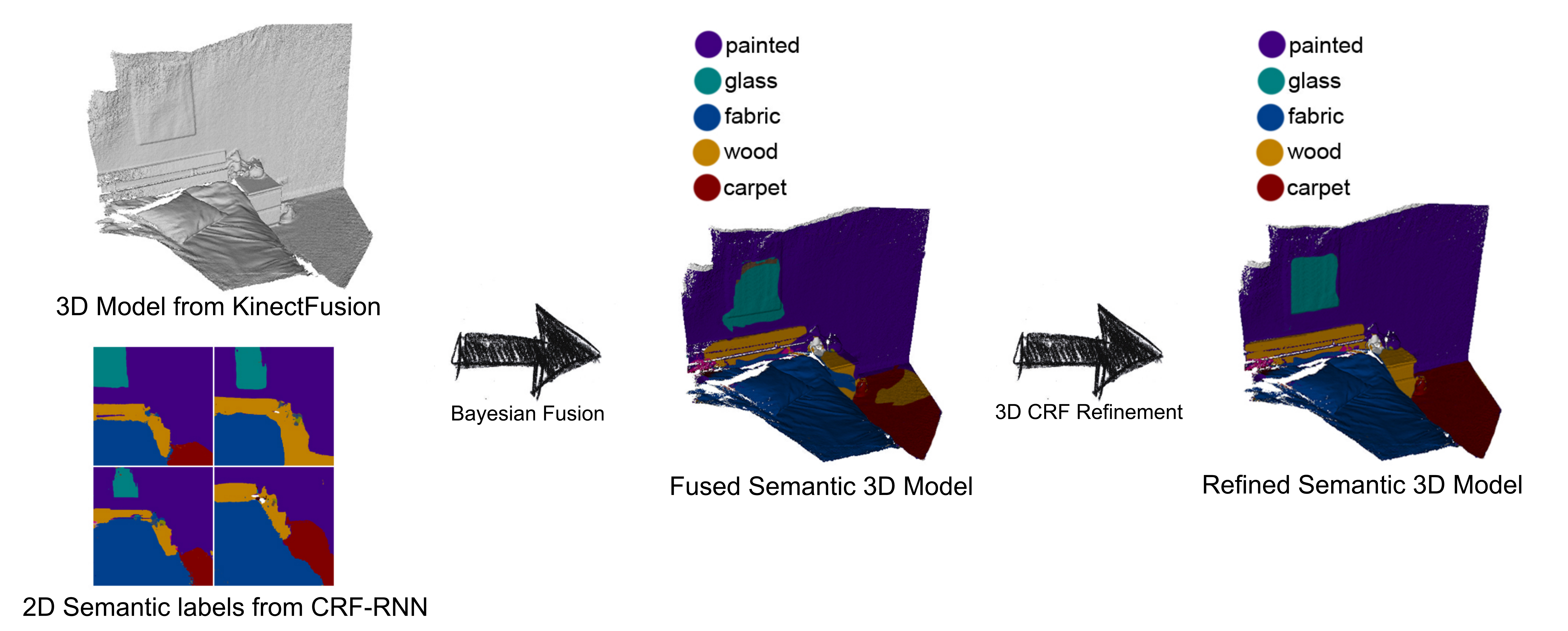}

\caption{3D semantic label fusion and refinement.}
\label{fuse}       
\end{figure*}

\begin{figure}[!t]
\centering
\subfloat[]{\includegraphics[width=0.23\textwidth]{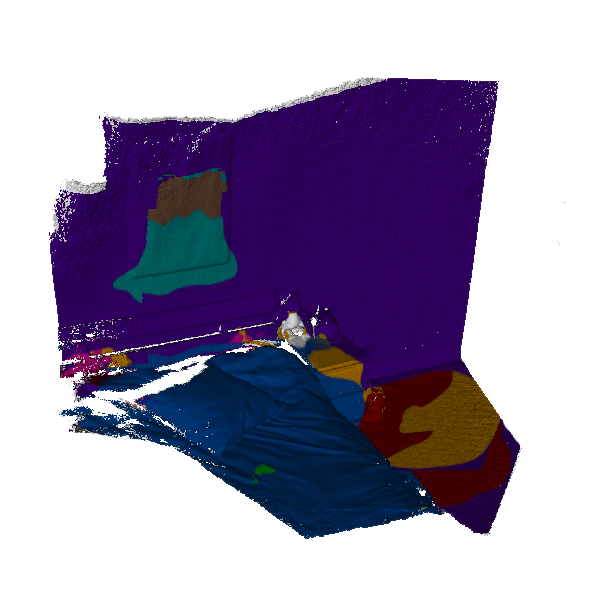}}\
\subfloat[]{\includegraphics[width=0.23\textwidth]{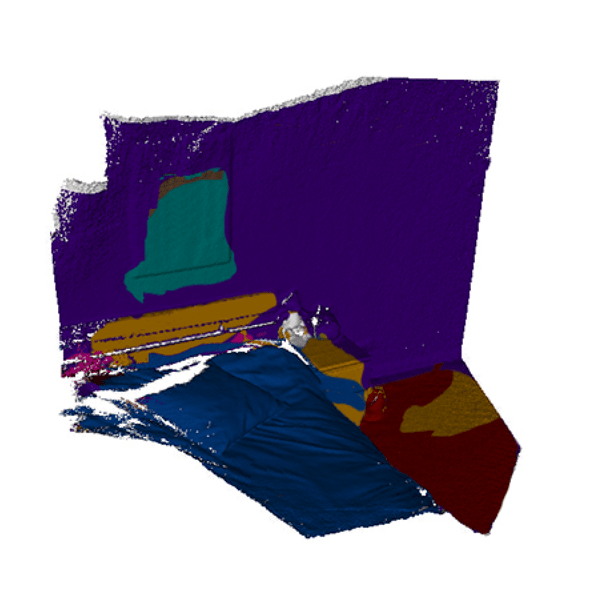}}\\
\caption{(a) 3D semantic label fusion using direct mapping. (b) 3D Semantic Label Fusion with Bayesian fusion.}
\label{nocrf}       
\end{figure}
The trained neural network for material recognition only infers object material properties in the 2D space. As the camera pose for each image frame is known, we can project the semantic labels onto the 3D model as textures. A direct mapping can cause information overlapping, since accumulated weak predictions and noises can lead a bad fusion result as shown in Figure \ref{nocrf} (a), where boundaries between different materials are blurred. We solve this issue by utilising the dense pixel-wise semantic probability distribution produced by the neural network over every class. Therefore, we are able to improve the fusion accuracy by projecting the labels with a statistical approach using the Bayesian fusion \cite{Armeni2016} \cite{Hermans2014} \cite{Zhao2017} \cite{McCormac2016}. Bayesian fusion enables us to update the label prediction $l_{i}$ on 2D images $I_{k}$ within the common coordinate frame of the 3D model.

\begin{equation}
P\left ( x=l_{i}|I_{1,...,k} \right )=\frac{1}{Z}P\left ( x=l_{i}|I_{1,...,k-1} \right ) P\left ( x=l_{i}|I_{k} \right )
\end{equation}

where Z is a constant for the distribution normalization. 
The label of each voxel is updated with the corresponding maximum probability $p\left ( x_{max}=l_{i}|I_{1,...,k} \right )$. The Bayesian fusion guides the label fusion process and ensures an accurate mapping result over the time to overcome the accumulated errors to some extent. Figure \ref{nocrf} (a) shows without the Bayesian fusion, the label fusion results are less clear due to the overlapping of weak predictions. In contrast, \ref{nocrf} (b) with the Bayesian fusion, the fusion results are much cleaner.

After semantic information fusion into the 3D model, we can get a 3D semantic labelled model. Although the Bayesian fusion is used to guide the fusion process, due to accumulation of the 2D segmentation error and the tracking error, in some area, the semantic information still not perfectly matches the model structure (see Figure \ref{fuse}). Next we explain how to further improve the fusion accuracy by proposing a new CRF label refinement process on 3D structures.

\subsection{3D Structural CRF Label Refinement}

We further improve the accuracy of the 3D labelling with a final refinement step on the semantic information using the structural and color information of the vertices of the 3D semantic model. From the fully connected CRF model, the energy of a label assignment $x$ can be represented as the sum of unary potentials and pairwise potentials over all $i$ pixels:

\begin{equation}
\label{crf}
    E\left ( x \right ) = \sum_{i}\psi _{u}\left ( x_{i} \right )+\sum_{i}\sum_{j\in N_{i}}\psi _{p}\left ( x_{i},x_{j} \right )
\end{equation}

where the unary potential $\psi _{u}\left ( x_{i} \right )$ is the cost (inverse likelihood) of the $i_{th}$ vertex assigning with the label x. In our model implementation, we use the final probability distribution from the previous Bayesian Fusion step as the unary potential for each label of every vertex. The pairwise potential is the energy term of assigning the label $x$ to both $i_{th}$ and $j_{th}$ vertices. We follow the efficient pairwise edge potentials in \cite{Kraehenbuehl2011} by defining the pairwise energy term as a linear combination of Gaussian kernels:

\begin{equation}
    \psi_{p}\left ( x_{i},x_{j} \right ) = \mu\left ( x_{i},x_{j} \right )\sum_{m=1}^{M}w^{(m)}k_{G}^{(m)}\left ( f_{i},f_{j} \right )
\end{equation}
where $w^{m}$ are the weights for different linear combinations, $k_{m}^{G}$ are $m$ different Gaussian kernels that $f_{i}$ and $f_{j}$ correspond to different feature vectors. Here, besides the commonly used feature space in \cite{Kraehenbuehl2011} \cite{crfasrnn_iccv2015} such as the color and the spatial location, the normal direction is also considered as a feature vector to take the full advantage of our 3D reconstruction step:

    \begin{align*}
        k_{G}\left ( f_{i},f_{j} \right ) = &w^{(1)}exp\left ( -\frac{\left | p_{i}-p_{j}\right |^{2} }{2\theta _{p}^{2}} \right )\\ + &w^{(2)}exp\left ( -\frac{\left | p_{i}-p_{j}\right |^{2} }{2\theta _{pI}^{2}}-\frac{\left | I_{i}-I_{j}\right |^{2} }{2\theta _{I}^{2}} \right )\\ + &w^{(3)}exp\left ( -\frac{\left | p_{i}-p_{j}\right |^{2} }{2\theta _{pn}^{2}} -\frac{\left | n_{i}-n_{j}\right |^{2} }{2\theta _{n}^{2}} \right )
    \end{align*}

where $p_{i}$ and $p_{j}$ are pairwise position vectors; $I_{i}$ and $I_{j}$ are pairwise RGB color vectors; $n_{i}$ and $n_{j}$ are pairwise normal directional vectors. The first term is the smoothness kernel assuming that the nearby vertices are more likely to be in the same label, which can efficiently remove small isolated regions~\cite{Shotton2009}\cite{Kraehenbuehl2011}. The second term represents the appearance kernel that takes into account of the color consistency, since the adjacent vertices with similar color(s) are more likely to have the same label. The third term is the surface kernel which utilizes the 3D surface normal as a feature that vertices with similar normal directions are more likely to be the same label.

By minimizing Equation \ref{crf}, semantic labels on our 3D model are further refined according to the color and the geometric information, which can efficiently eliminates the ''label leaking'' problem caused by the 2D semantic segmentation errors and the camera tracking errors (see Figure \ref{fuse}). 

\subsection{Interaction Interface}


A user interface is developed with two layers. The top layer displays the current video stream from a RGB-D camera, whilst the semantic 3D model serves as a hidden physical interaction layer to provide an interaction interface. In the interactive MR application, a virtual camera is synchronized with predicted camera poses for projecting the 3D semantic model onto the corresponding view of the video stream. Figure \ref{eval} shows that the back layer of the interface displays the video stream feed from the RGB-D camera; A semantic interaction 3D model is in front of the video layer for handling interactions of different materials (Green: glass, Purple: painted, Blue: fabric, Yellow: wood, Red: carpet). The virtual and real physical interactions are performed on the interaction model. The context-aware interaction model is invisible to allow users interact with the real-world objects to experience an immersive MR environment. The interaction layer also computes real-time shadows to make the MR experience even more realistic. An oct-tree data structure accelerates the ray-casting queries for the material properties to improve the real-time performance. Finally, corresponding physical interactions based on semantic information e.g. different materials are achieved through physics simulations. 


\section{Example Applications}
\label{applications}
\begin{figure}[!]
\centering
\subfloat[]{\includegraphics[width=0.23\textwidth]{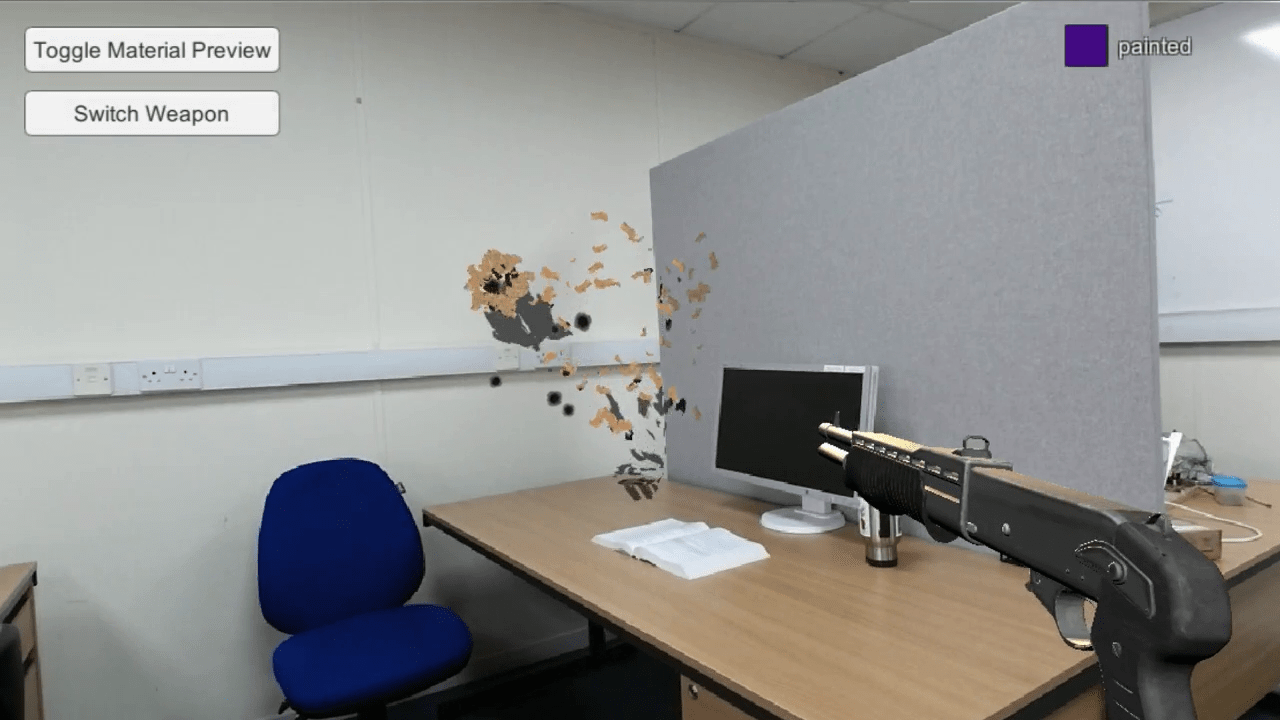}}\
\subfloat[]{\includegraphics[width=0.23\textwidth]{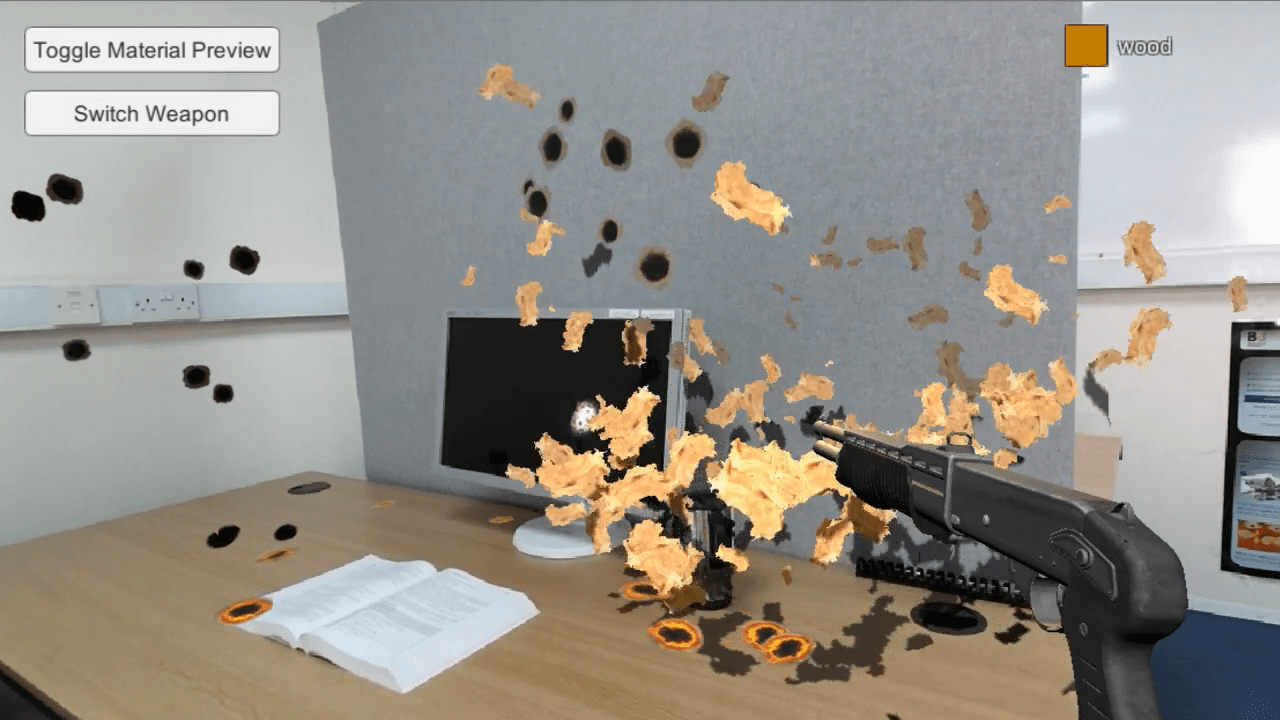}}\\
\subfloat[]{\includegraphics[width=0.23\textwidth]{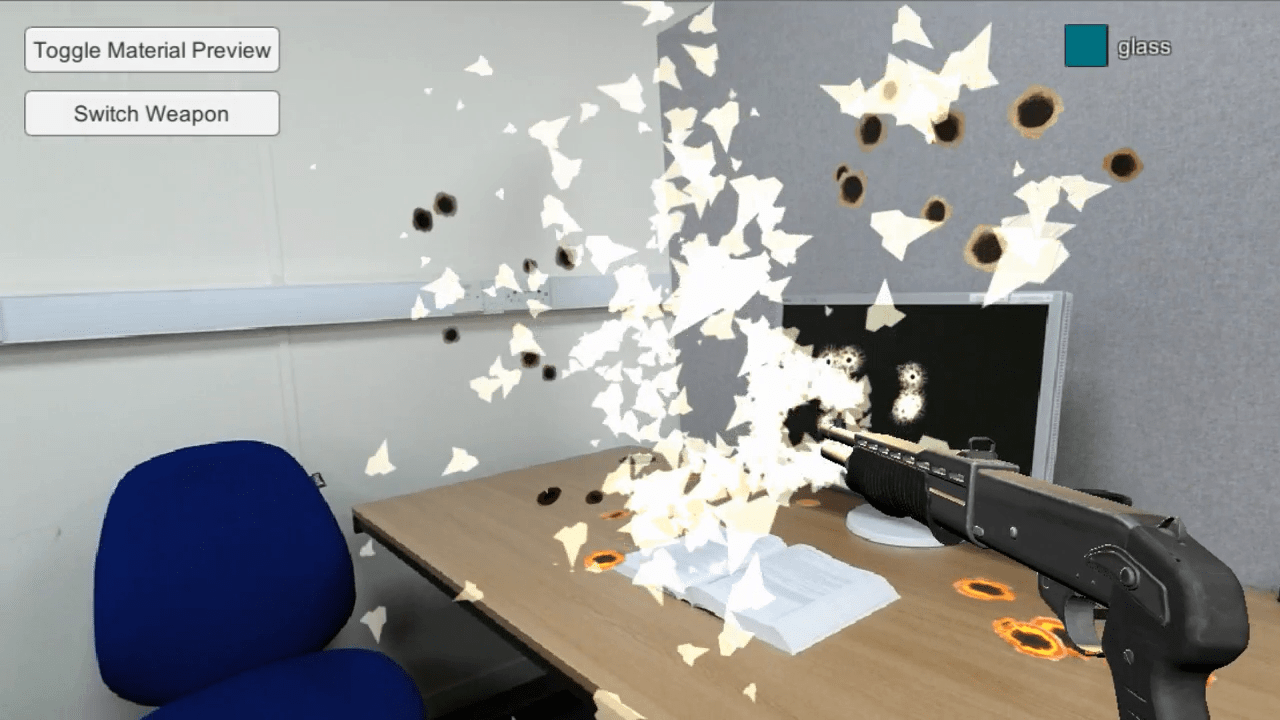}}\
\subfloat[]{\includegraphics[width=0.23\textwidth]{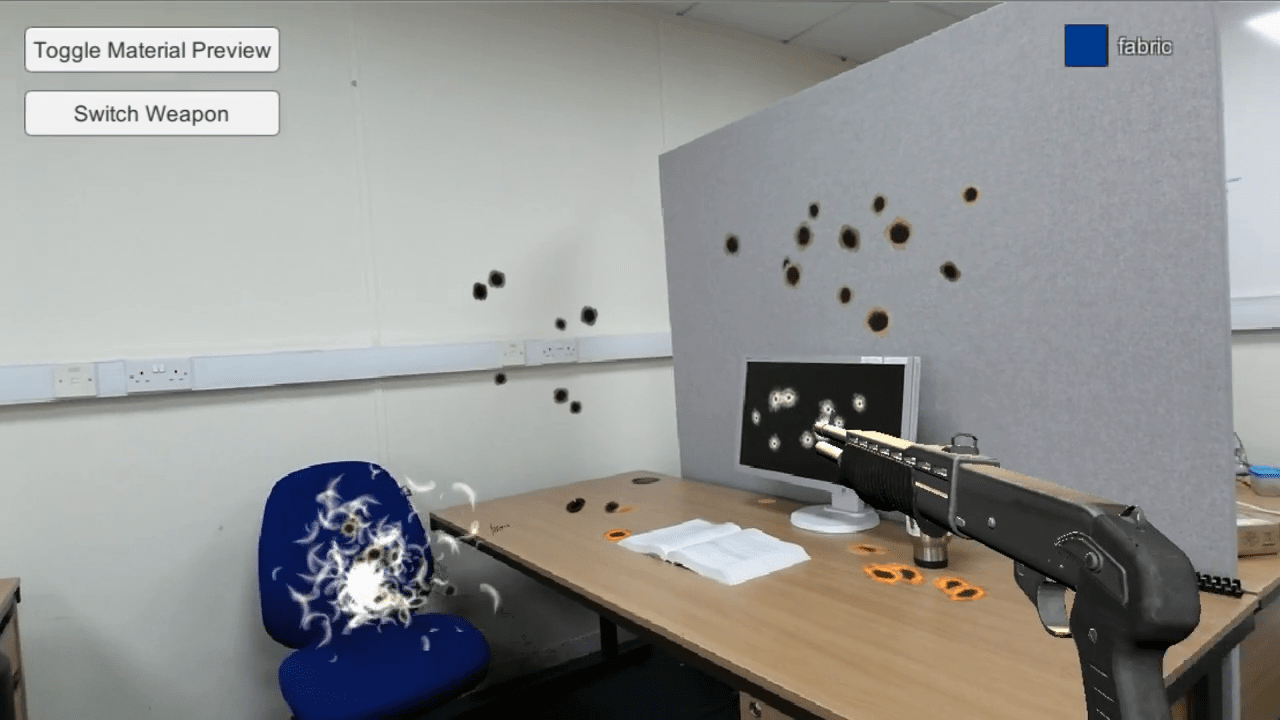}}\\
\caption{The screenshots of our MR shooting game. The interaction is different when shooting (a)walls, (b)desks, (c)computer screen and (d)chair.}
\label{shooting}       
\end{figure}

Based on our implementation, two FPS games are developed to demonstrate the concept of the proposed material-aware interactive MR. Next, we describe the design of the interactions and evaluations. 

Games are interaction demanding applications driven by computational performance and accurate interactions in virtual space. We designed two MR game that can directly interact with the real-world objects. A shooting game is designed to show material-aware interactions between bullets and the real world objects. The shoot scenario is chosen, because we want to test the accuracy of the semantic 3D model using ray-cast queries. In this game, as shown in Figure \ref{shooting}, multiple interactions for different materials have been implemented including different bullet holes, flying chips and hitting sound when hitting different objects: (a)walls, (b)desks, (c)computer screen and (d)chair. The interaction for different material context is as real as possible.

Another way to show the capability of the context-aware framework is to match the interaction results to the user's anticipation of the interaction results using everyday scenarios that familiar to users, i.e. testing the immersive experience of the MR system from the user's perspective. The second example is designed to match the user expectations for material-specific physical interactions. 

\begin{figure}[]
\centering
\subfloat[]{\includegraphics[width=0.23\textwidth]{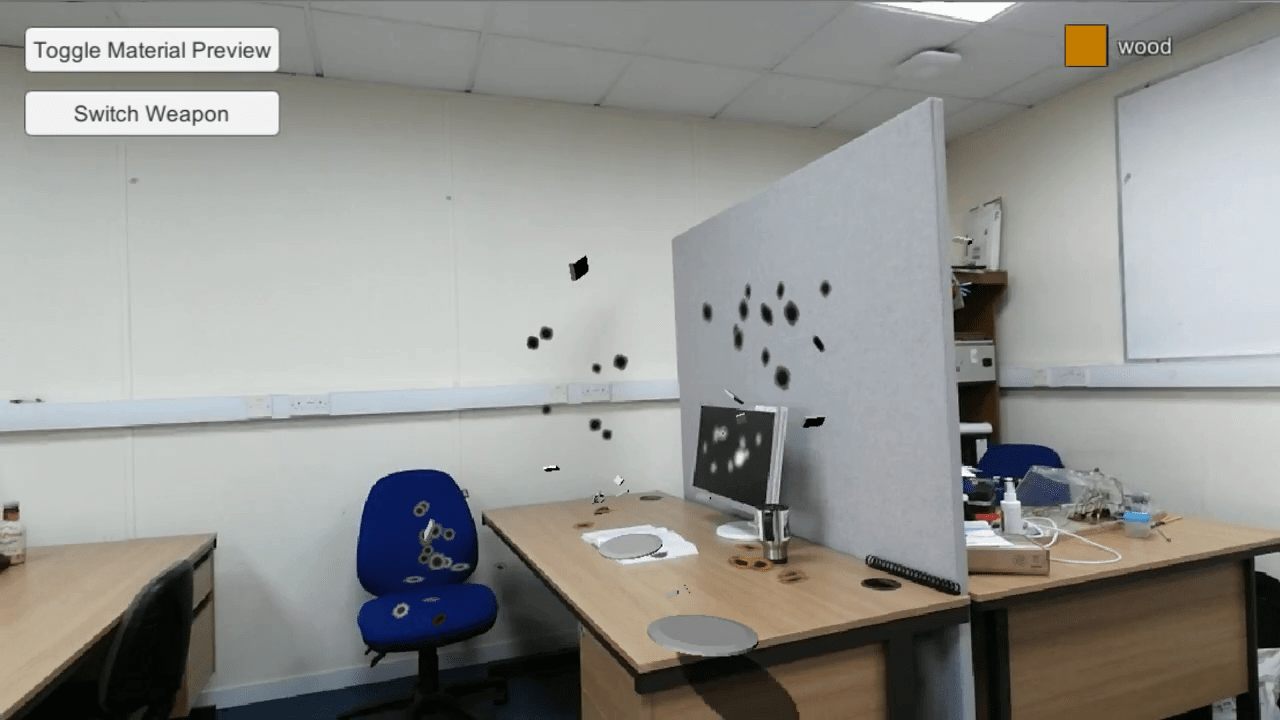}}\
\subfloat[]{\includegraphics[width=0.23\textwidth]{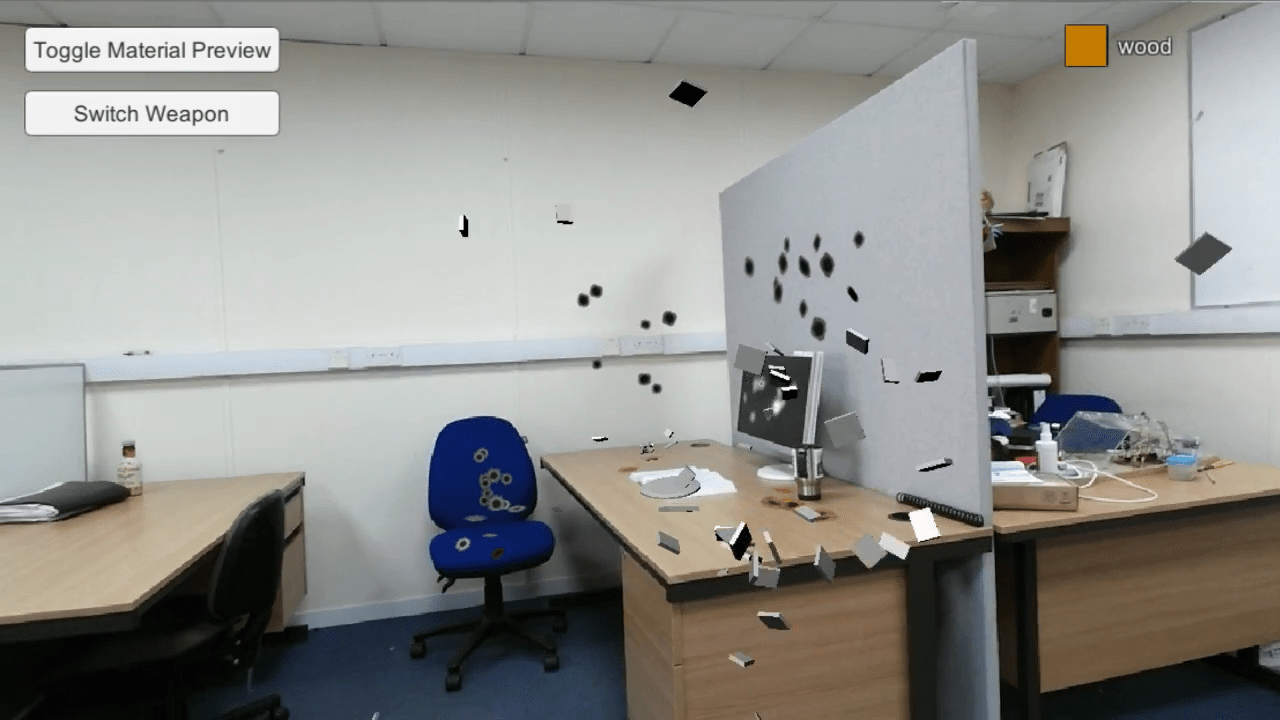}}\\
\subfloat[]{\includegraphics[width=0.23\textwidth]{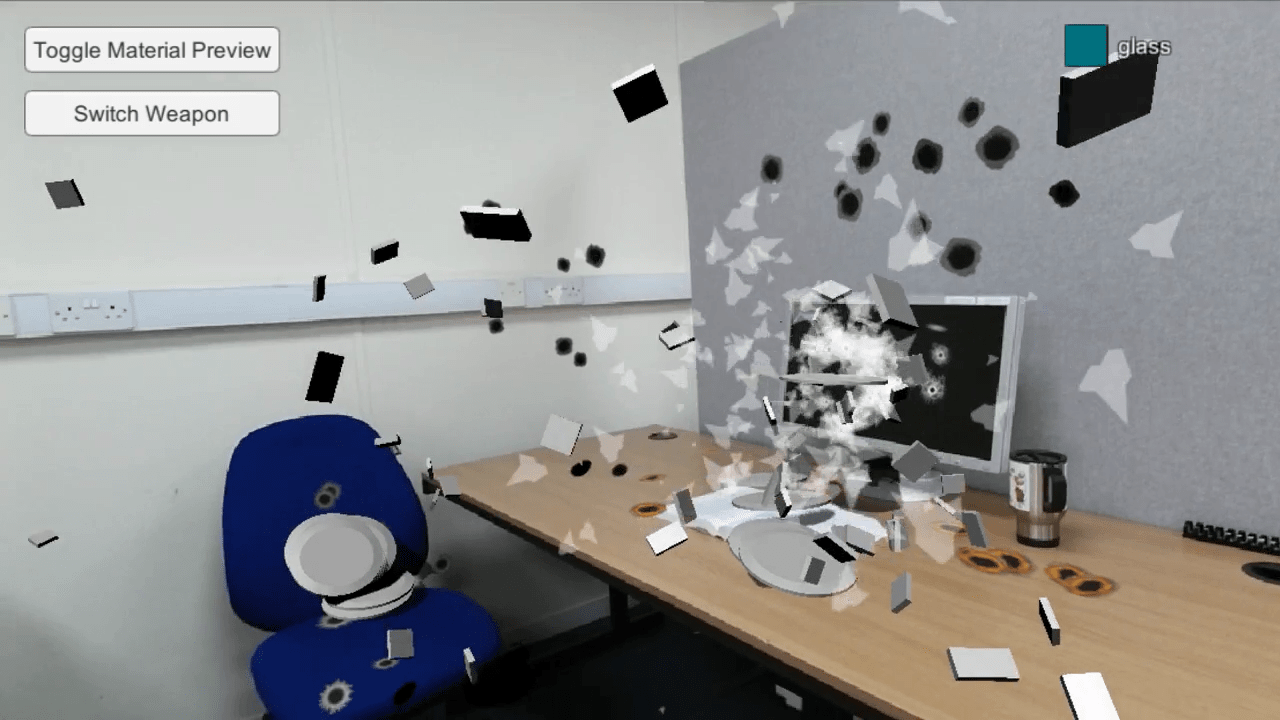}}\
\subfloat[]{\includegraphics[width=0.23\textwidth]{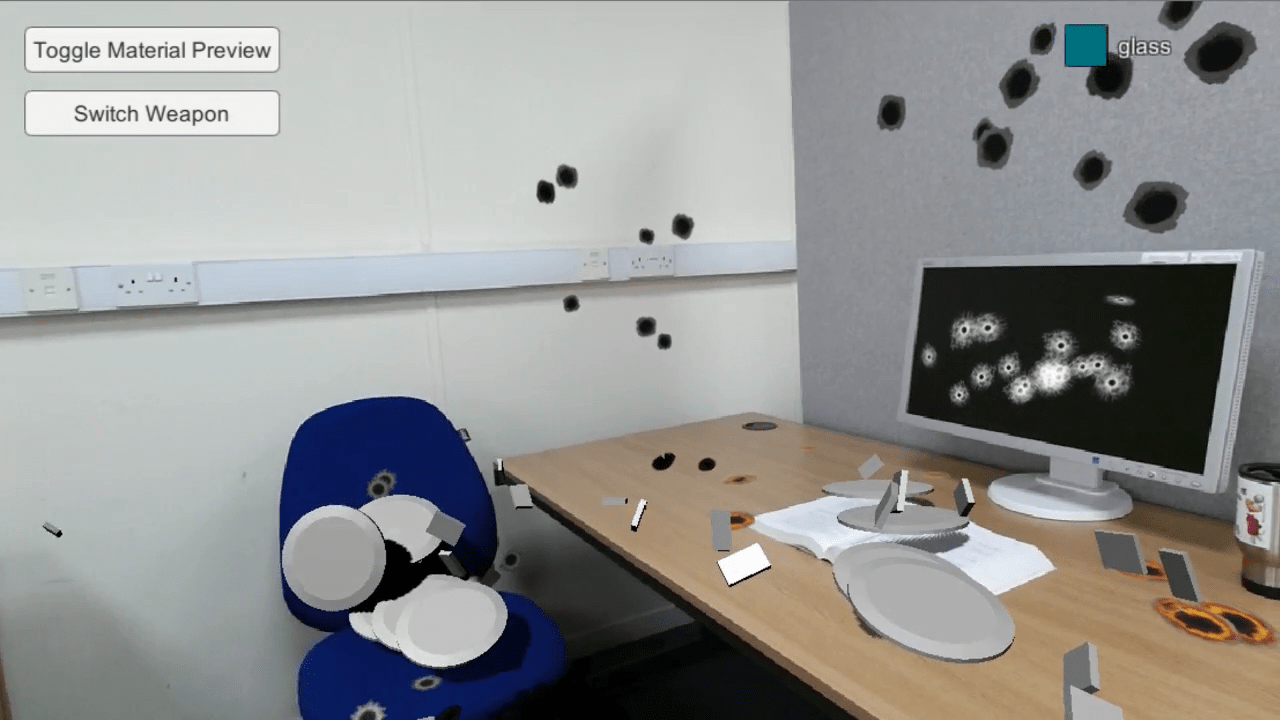}}\\
\caption{The screenshots of our MR throwing plates game. The interaction is different when throwing plates to (a)book, (b)desks, (c)computer screen and (d)chair.}
\label{throwing}       
\end{figure}

\begin{figure*}[!t]
\centering
\includegraphics[width=0.98\textwidth]{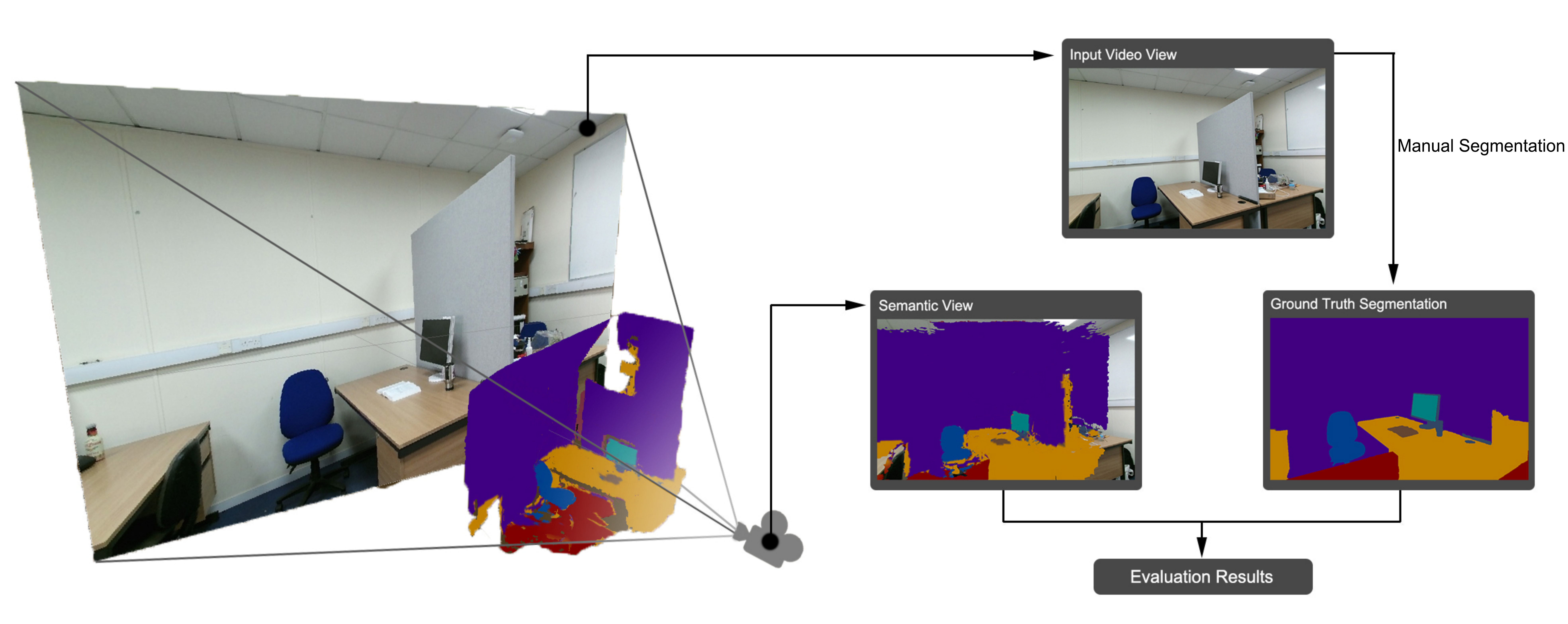}
\caption{The evaluation framework}
\label{eval}       
\end{figure*}

As shown in Figure \ref{throwing}, users throw virtual plates onto real world objects of the MR environment, resulting material-aware physical interactions induced by various material properties of the real objects. In Figures \ref{throwing} (a) and (b), virtual plates are broken when felling onto the desk, but bounced back when colliding with a book; in (c) when colliding with a computer screen, the plate is broken with the flying glass chips; in (d), the plate remain intact colliding with a soft chair.

\section{EXPERIMENTATION}

{\setlength{\tabcolsep}{0.3em}
\begin{figure*}[]
  \centering
  \begin{tabular}[c]{cccccc}
  FCN & CRF\textendash RNN & Ours\textendash noCRF & Ours & Ground Truth & Image \\
  \includegraphics[width=0.15\textwidth]{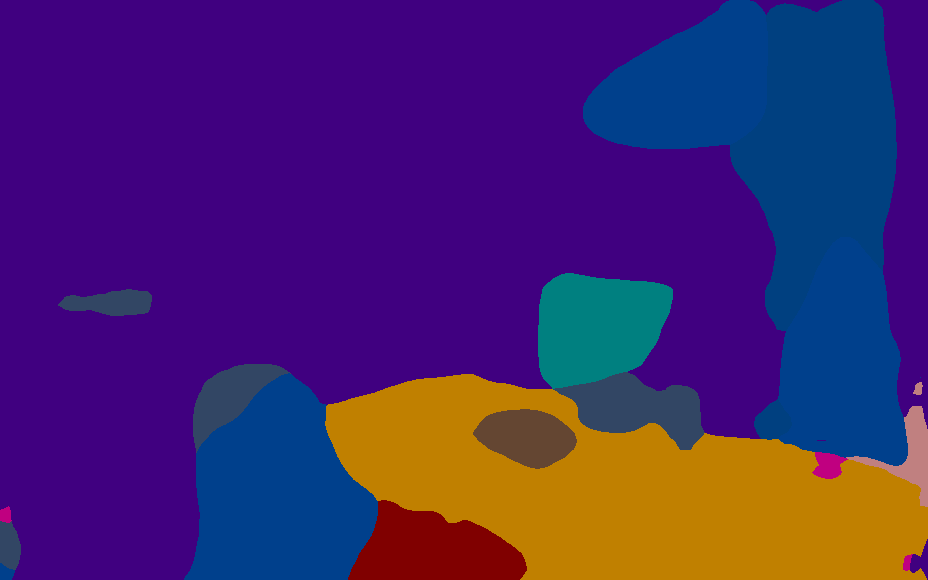} &
  \includegraphics[width=0.15\textwidth]{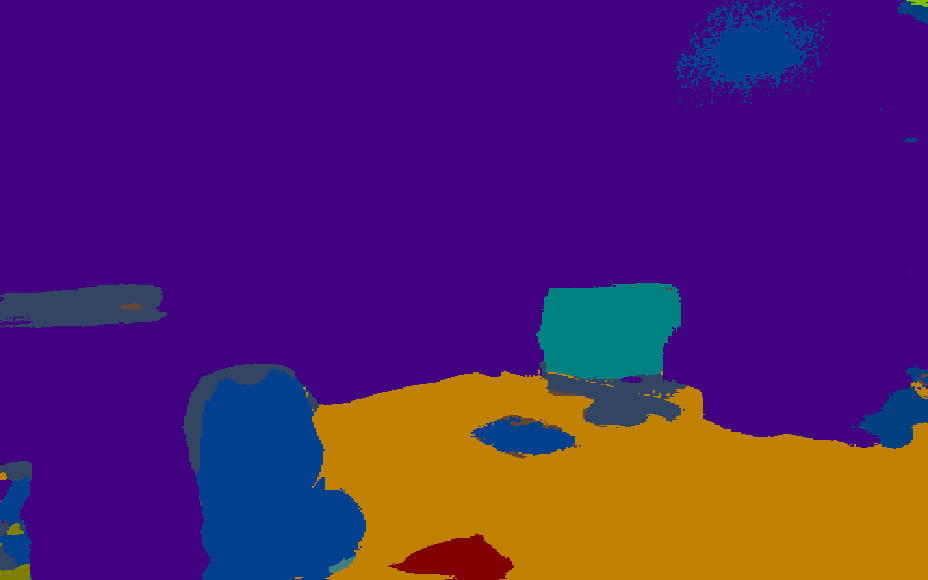} &
  \includegraphics[width=0.15\textwidth]{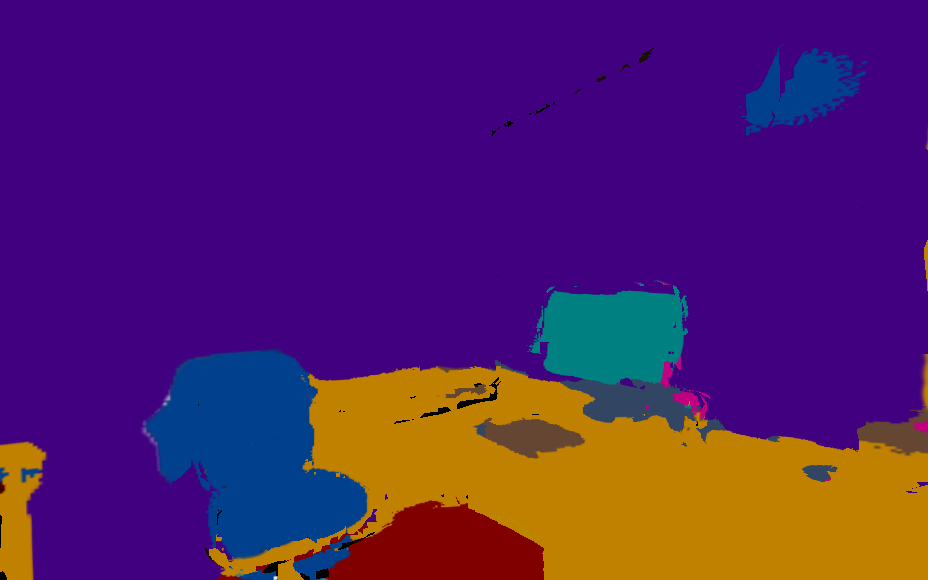} &
  \includegraphics[width=0.15\textwidth]{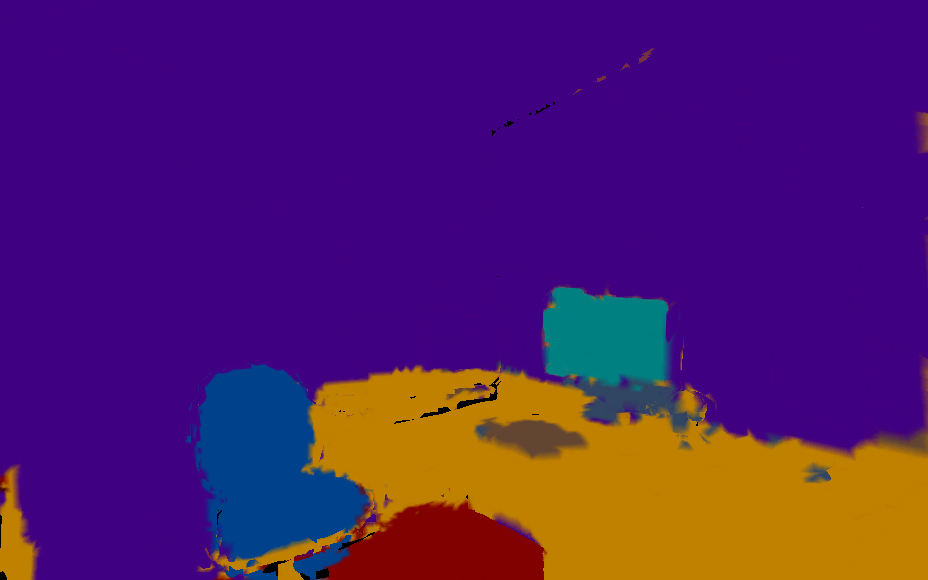} &
  \includegraphics[width=0.15\textwidth]{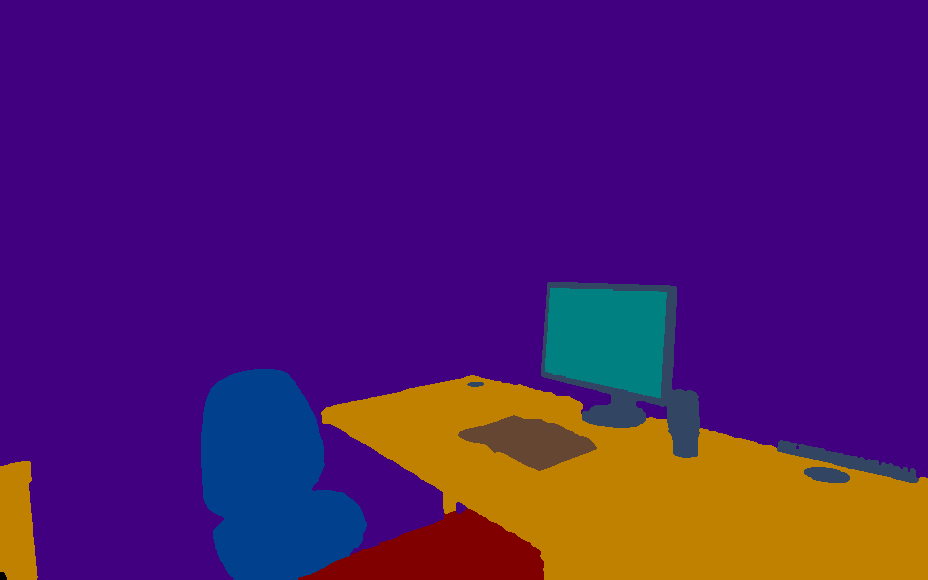} &
  \includegraphics[width=0.15\textwidth]{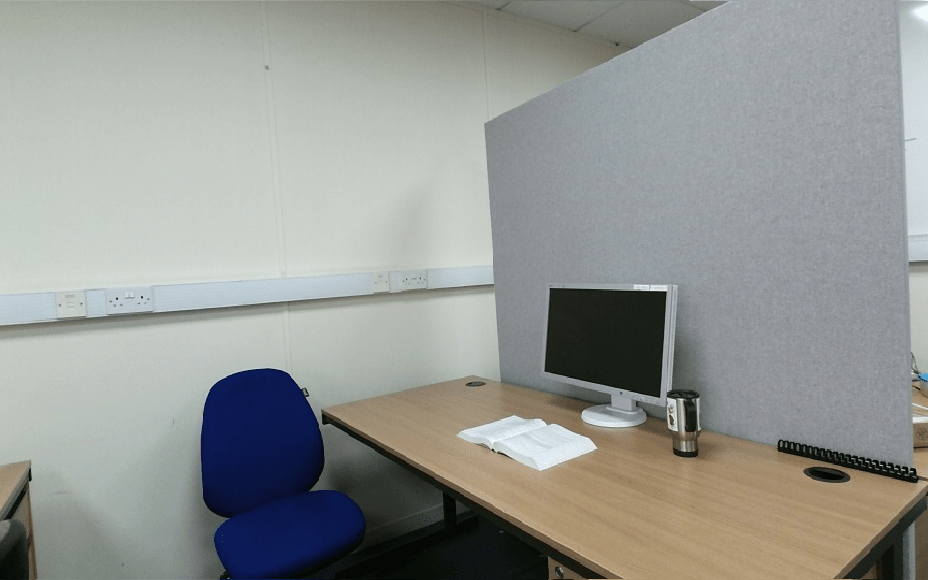} \\  \includegraphics[width=0.15\textwidth]{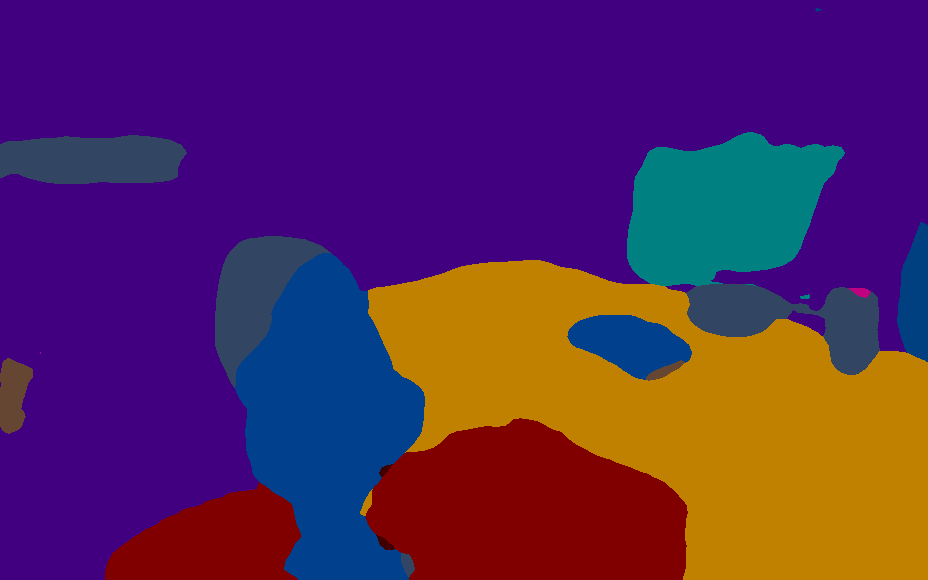} &
  \includegraphics[width=0.15\textwidth]{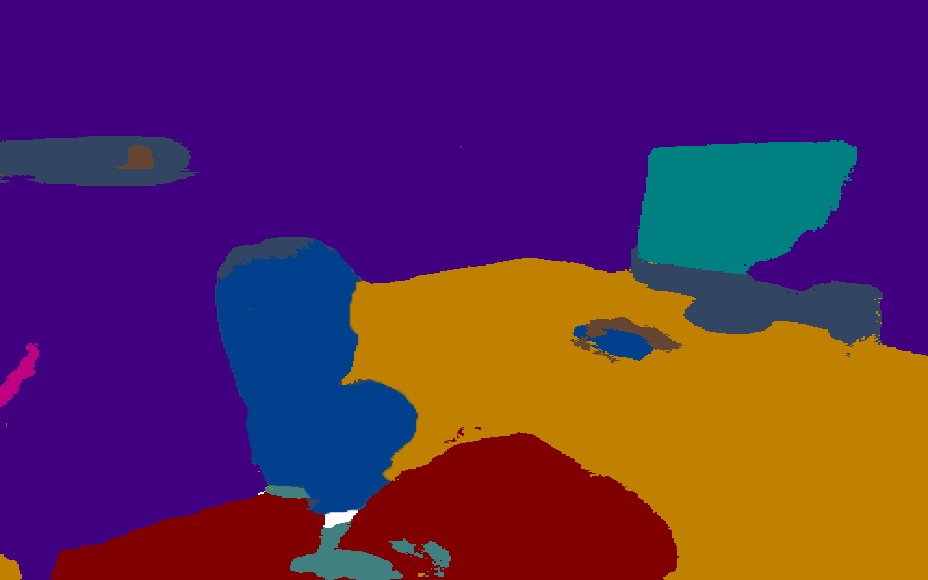} &
  \includegraphics[width=0.15\textwidth]{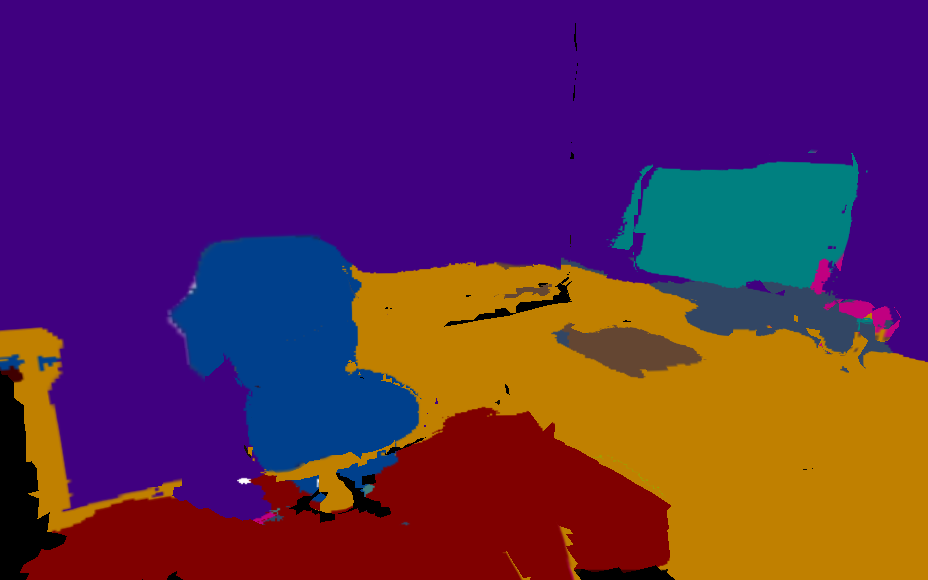} &
  \includegraphics[width=0.15\textwidth]{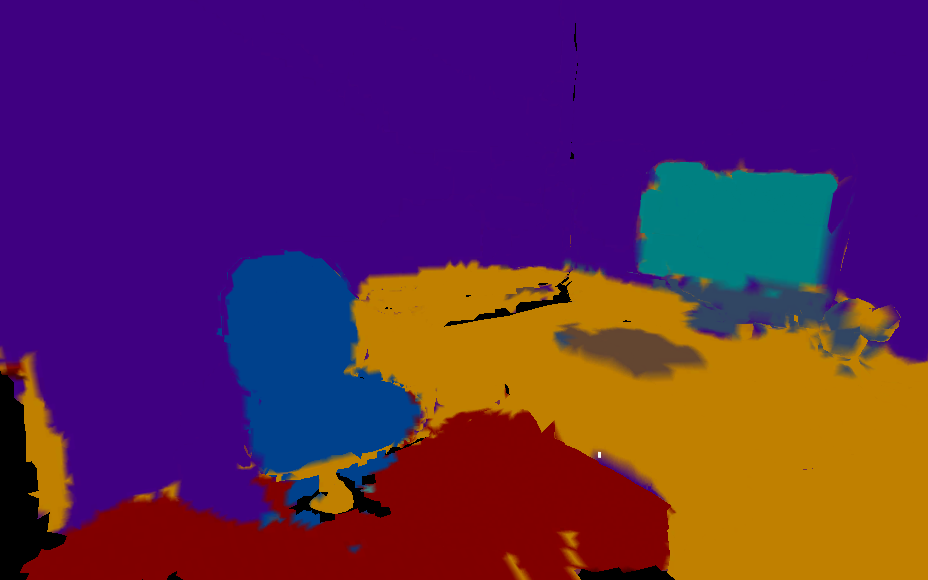} &
  \includegraphics[width=0.15\textwidth]{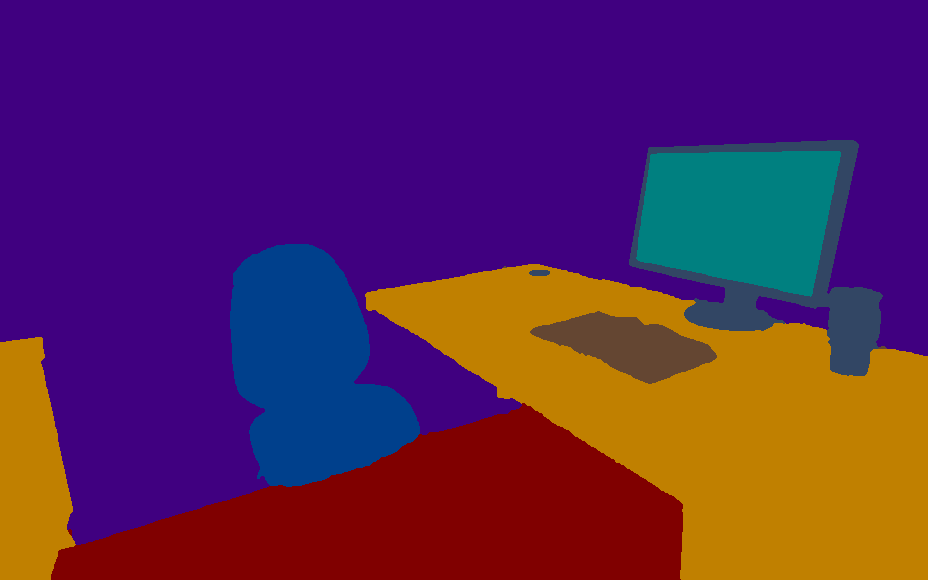} &
  \includegraphics[width=0.15\textwidth]{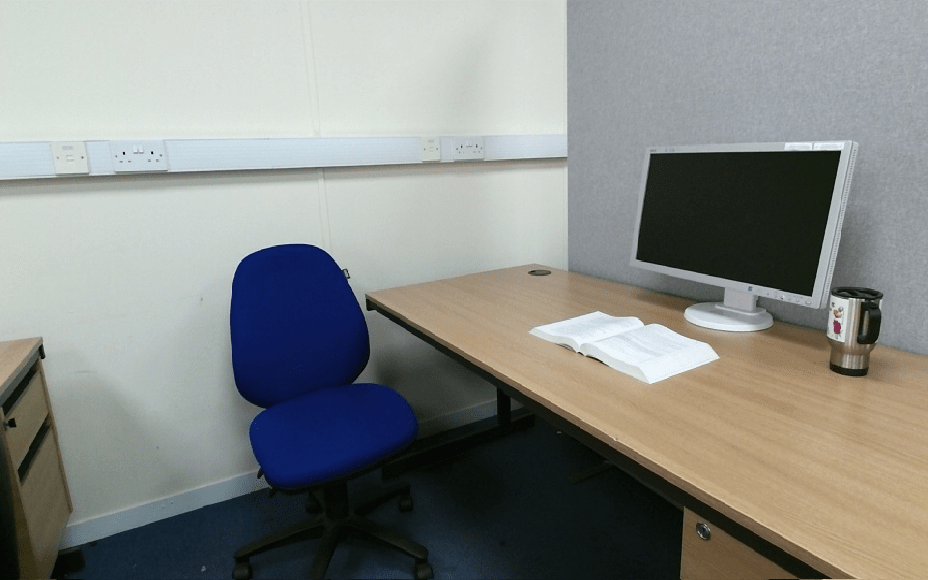} \\  \includegraphics[width=0.15\textwidth]{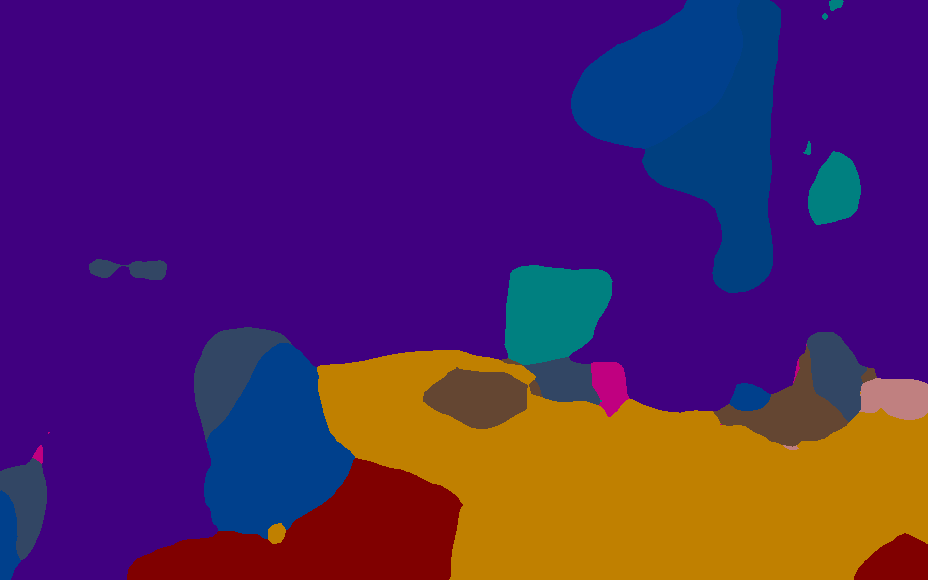} &
  \includegraphics[width=0.15\textwidth]{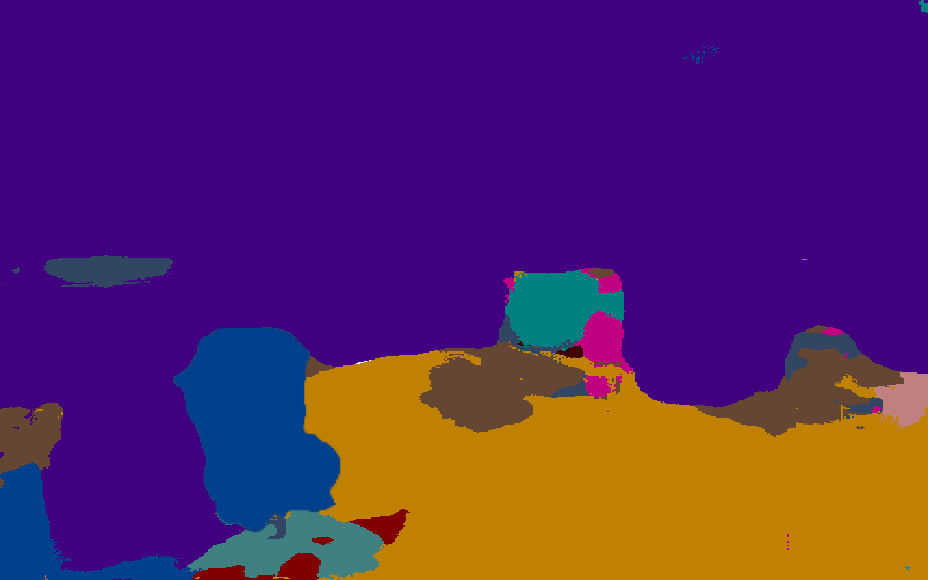} &
  \includegraphics[width=0.15\textwidth]{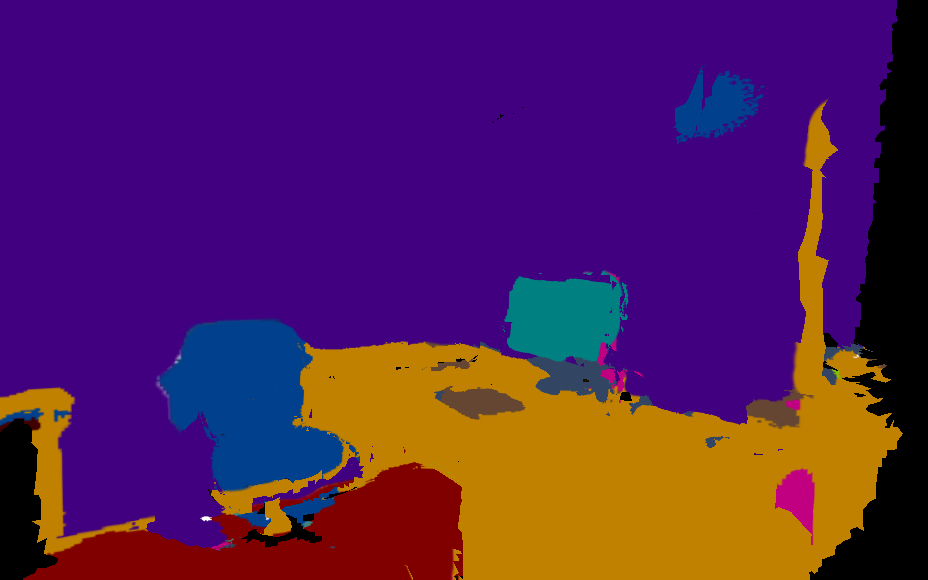} &
  \includegraphics[width=0.15\textwidth]{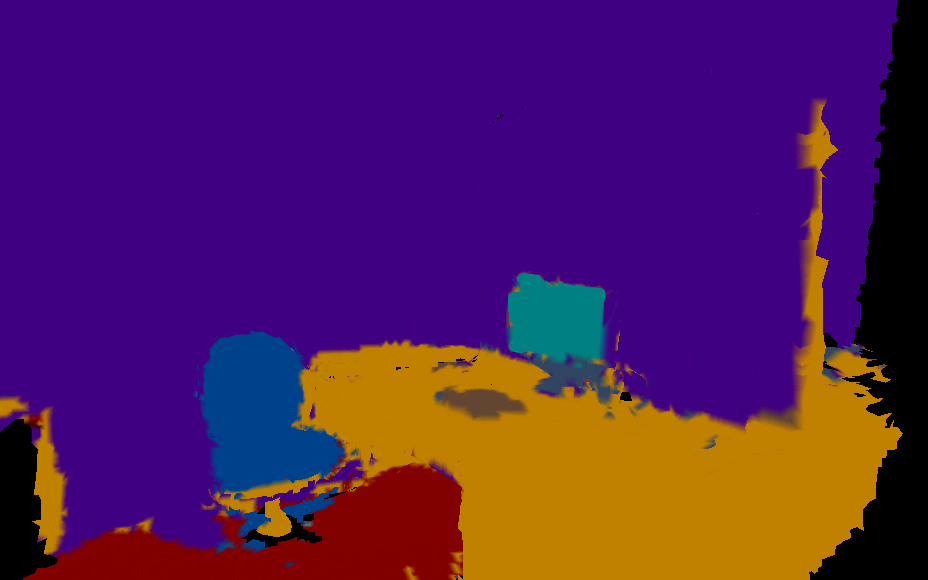} &
  \includegraphics[width=0.15\textwidth]{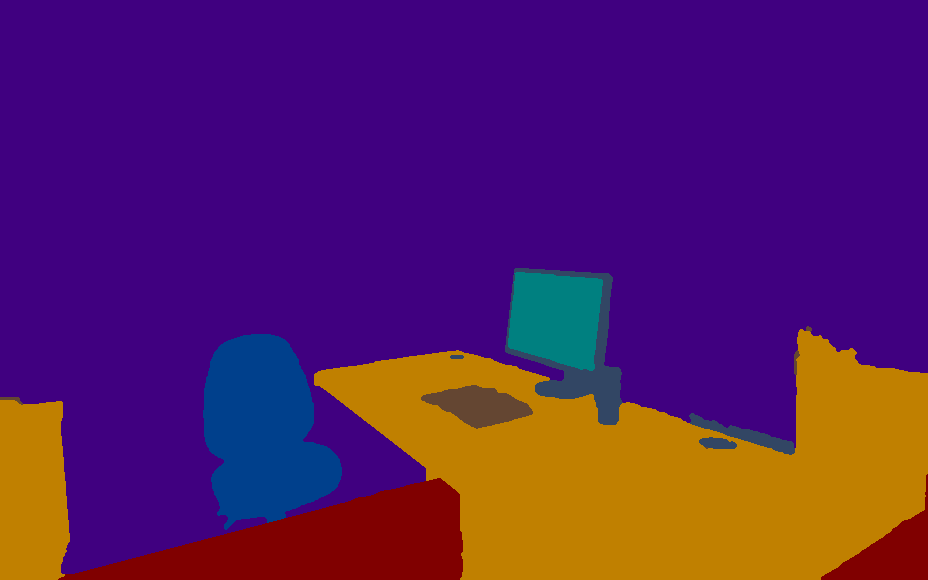} &
  \includegraphics[width=0.15\textwidth]{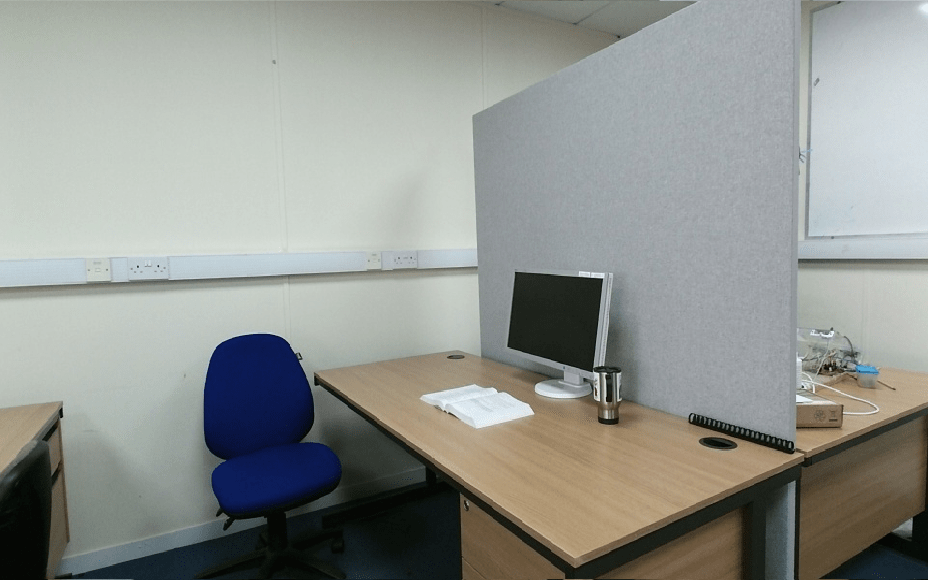} \\ 
  \includegraphics[width=0.15\textwidth]{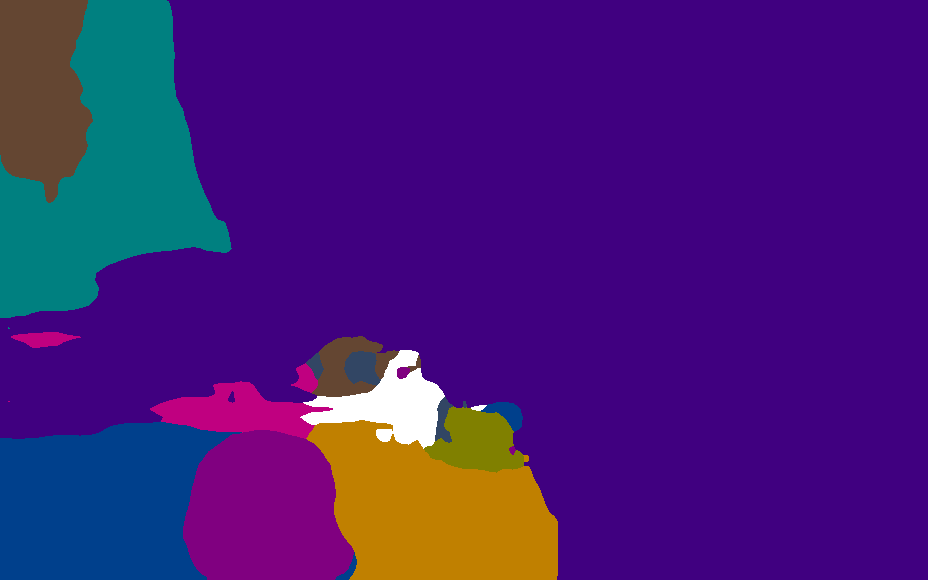} &
  \includegraphics[width=0.15\textwidth]{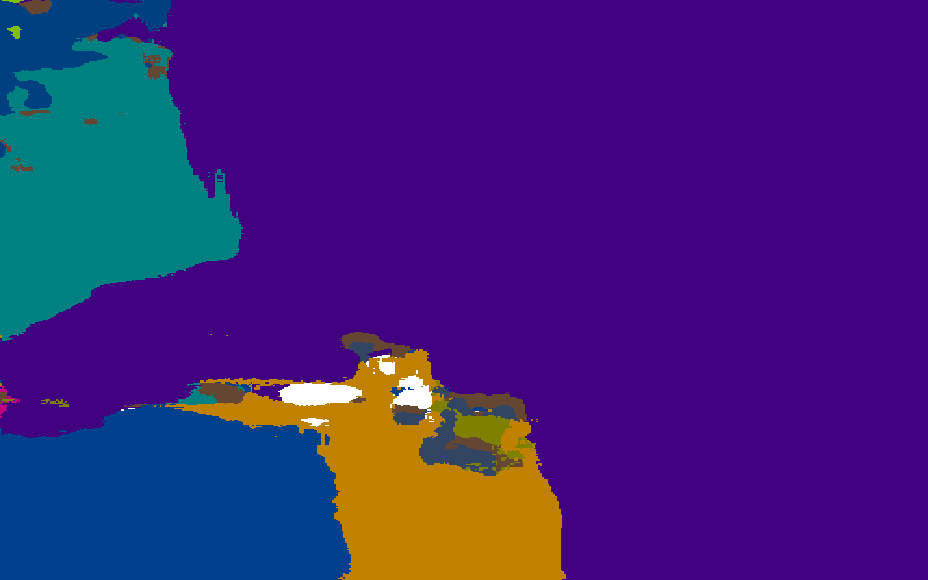} &
  \includegraphics[width=0.15\textwidth]{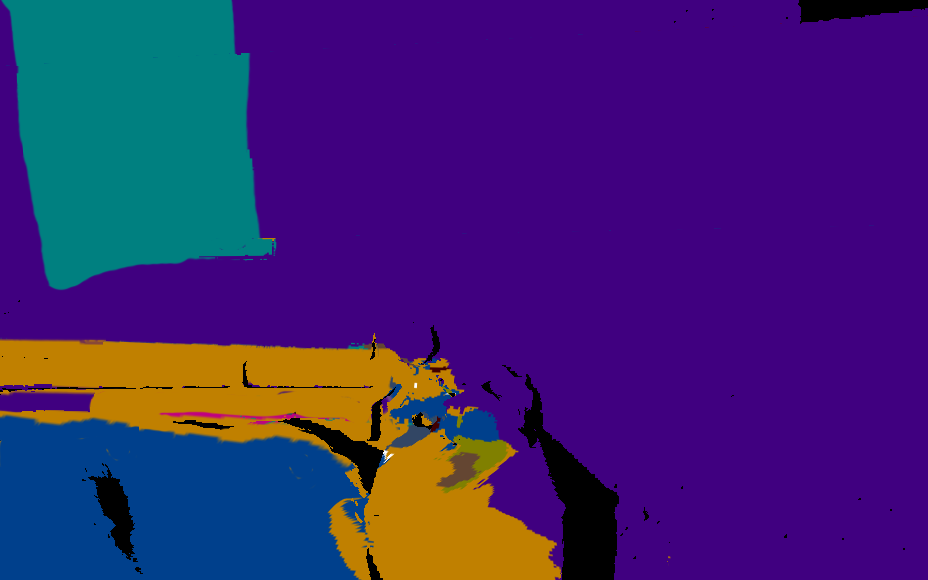} &
  \includegraphics[width=0.15\textwidth]{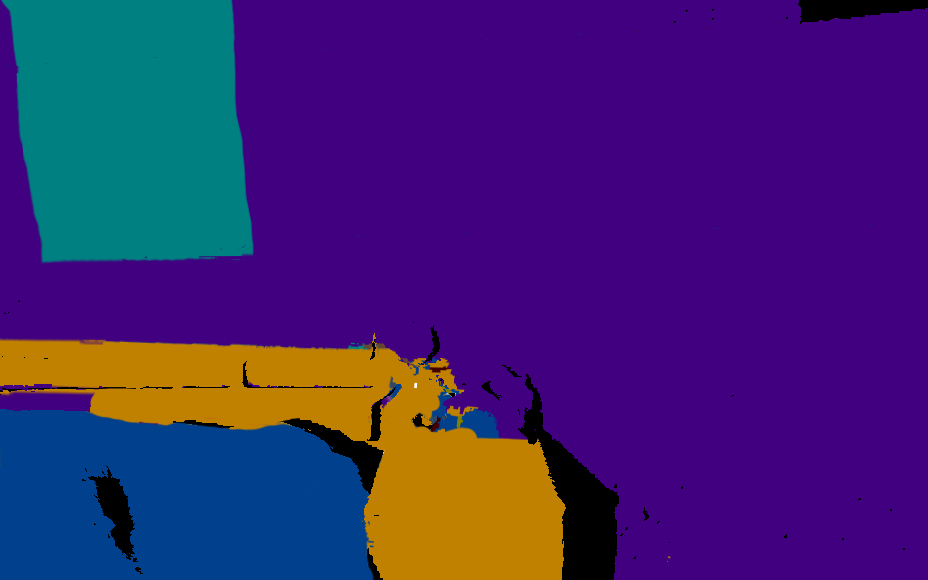} &
  \includegraphics[width=0.15\textwidth]{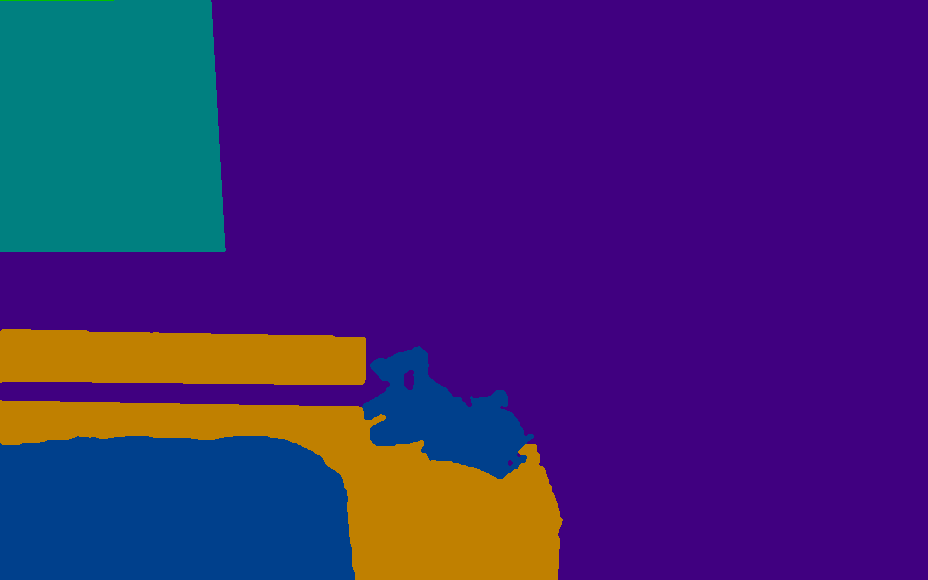} &
  \includegraphics[width=0.15\textwidth]{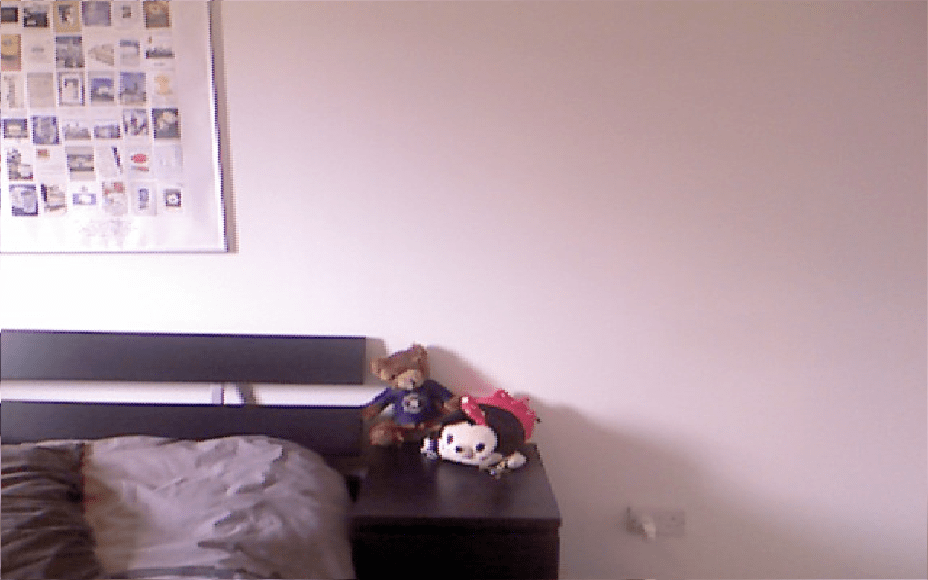} \\
  \includegraphics[width=0.15\textwidth]{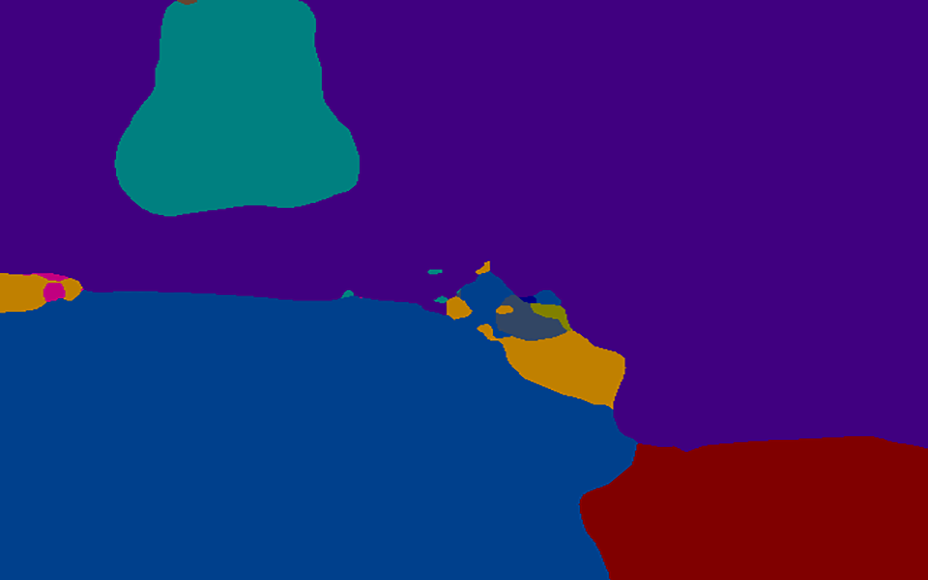} &
  \includegraphics[width=0.15\textwidth]{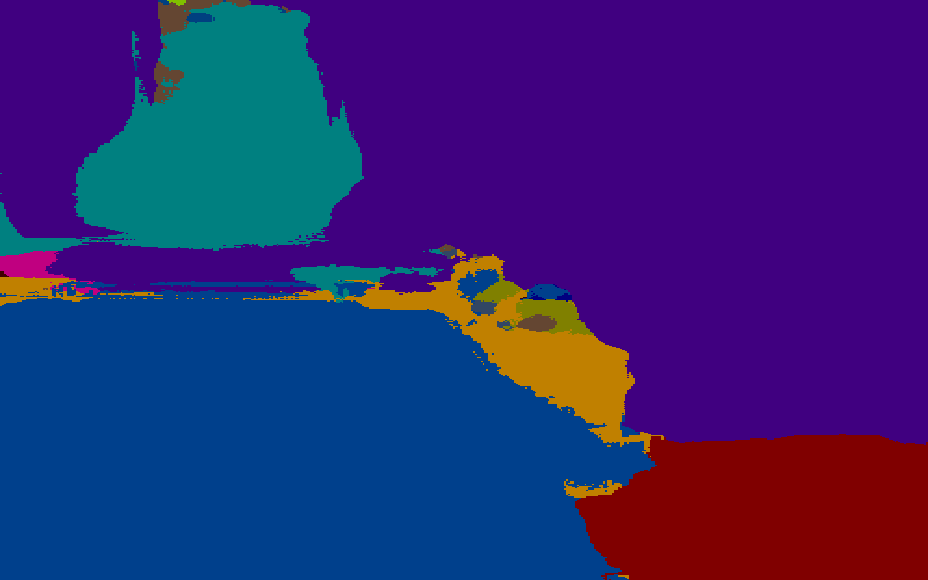} &
  \includegraphics[width=0.15\textwidth]{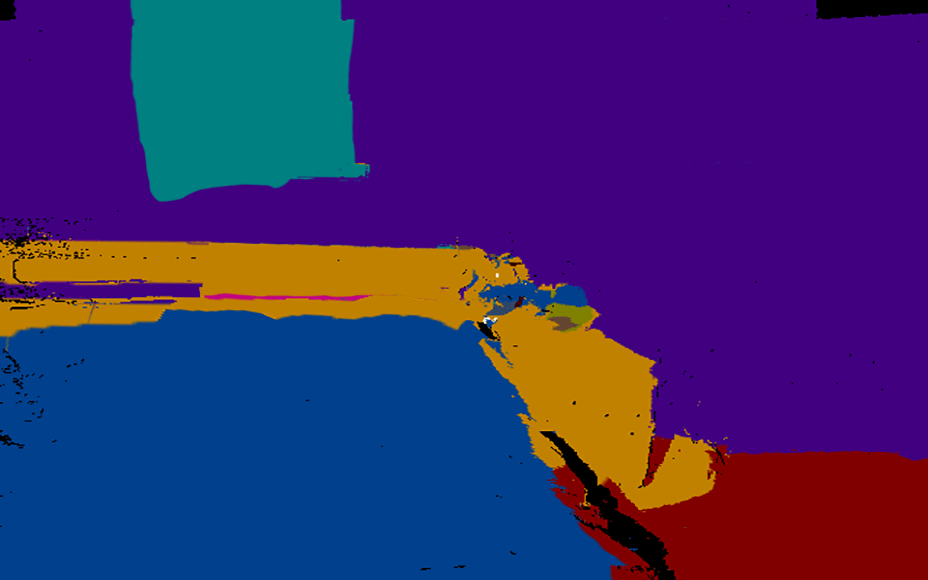} &
  \includegraphics[width=0.15\textwidth]{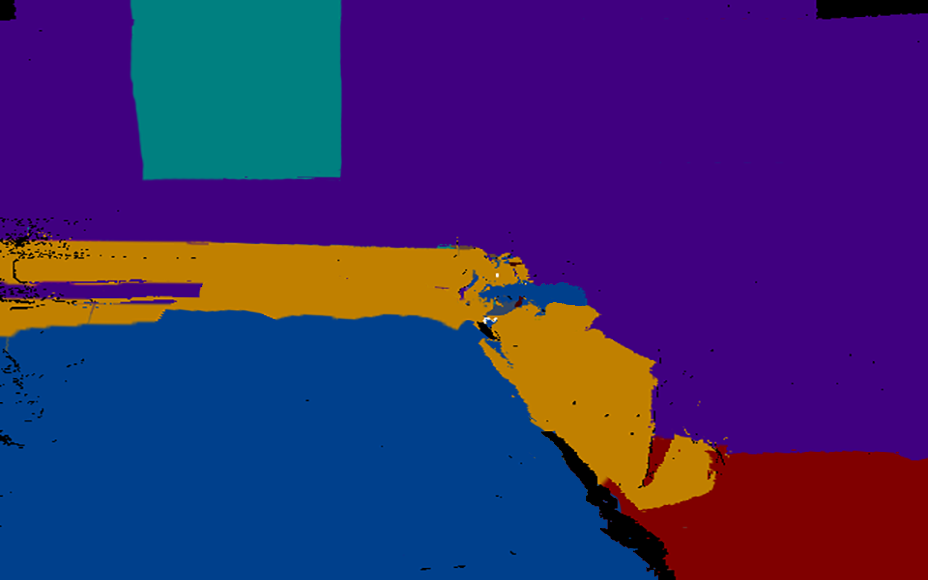} &
  \includegraphics[width=0.15\textwidth]{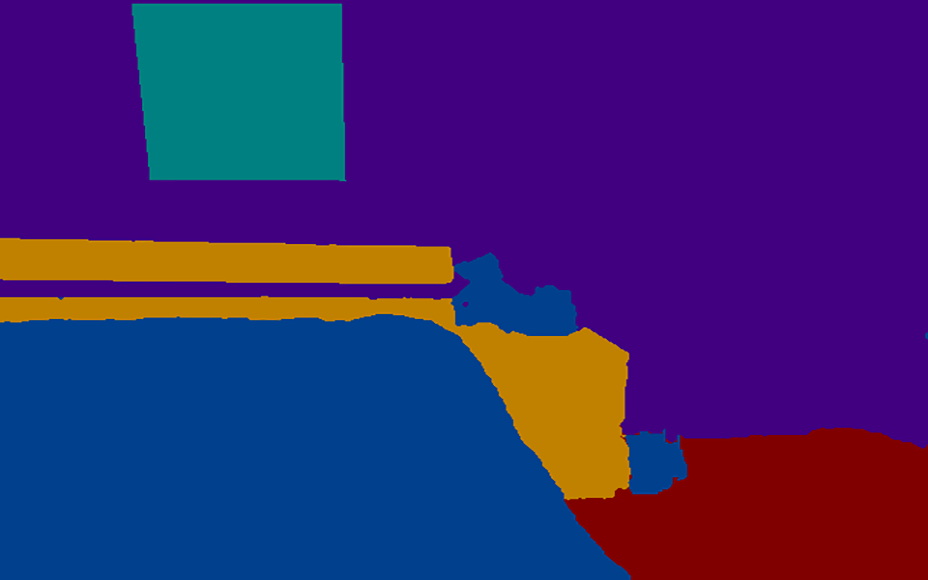} &
  \includegraphics[width=0.15\textwidth]{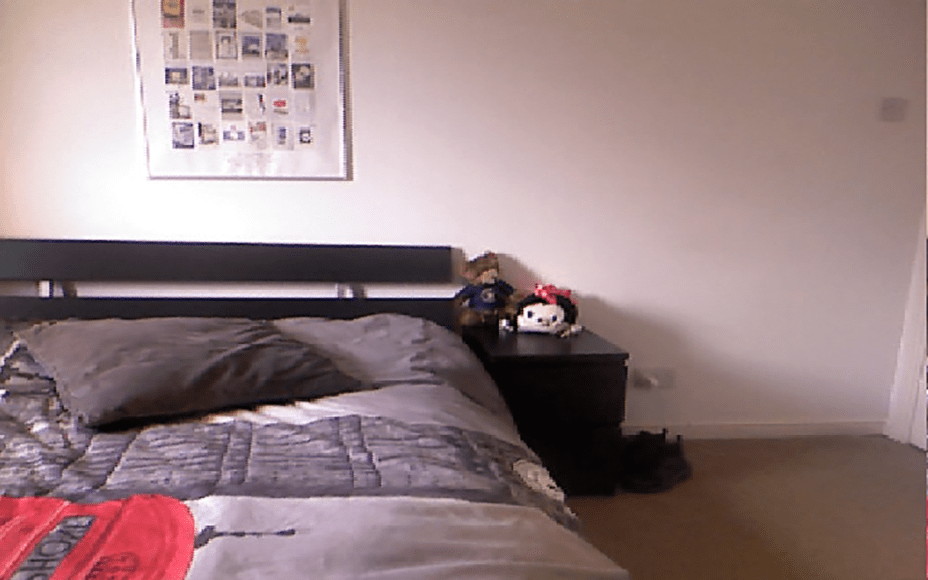} \\
  \end{tabular}
  \caption{Semantic segmentation samples for each column from left to right: FCN, CRF-RNN, Ours without CRF refinement, Ours with CRF refinement, Ground Truth, Input image}
  \label{compare} 
\end{figure*}}

\subsection{Accuracy Study}

Multiple factors affect the accuracy of the system: (1) the camera tracking, (2) the 3D model reconstruction, (3) the deep semantic material segmentation, (4) the 2D to 3D semantic model fusion and (5) the implementation of ray-casting. Therefore, it is extremely difficult to evaluate the accuracy of every single part of the system, separately. We conducted an end-to-end accuracy study to directly evaluate the accuracy of the dense ray-casting queries of the 3D semantic model, because it directly determines the accuracy of interactions. A total of 25 key-frames from two different scenes (office and bedroom) are selected, and at the same time, the 2D projections of the 3D semantic models are captured as the dense ray-casting query results at the corresponding key-frames (see Figure \ref{eva}). Ground truth for the accuracy evaluation is obtained by manually labelling 25 RGB images with the same material labels. The four common evaluation criteria \cite{Shelhamer2017} \cite{crfasrnn_iccv2015} for semantic segmentation and scene parsing evaluations are used to evaluate the variations of pixel accuracy and region intersection over union (IU).
\begin{enumerate}
    \setlength\itemsep{0.8em}
    \item pixel accuracy $\frac{\sum_{i}n_{ii}}{\sum_{i}t_{i}}$
    \item mean accuracy $\frac{1}{n_{c1}}\sum_{i}\frac{n_{ii}}{t_{i}}$
    \item mean IU $\frac{1}{n_{c1}}\sum_{i}\frac{n_{ii}}{t_{i}+\sum_{j}n_{ji}-n_{ii}}$
    \item frequency weighted IU $\frac{1}{\sum_{k}t_{k}}\sum_{i}\frac{t_{i}n_{ii}}{t_{i}+\sum_{j}n_{ji}-n_{ii}}$
\end{enumerate}
 where $n_{ij}$ represents the the number of pixels of class $i$ predicted to be class $j$; $n_{c1}$ is the total number of classes; $t_{i}=\sum_{j}n_{ij}$ is the total number of pixels of class $i$.

As can be seen from Table \ref{resultt}, after 2D-3D fusion, 3D refinement and finally 3D-2D projections, our framework can provide more accurate semantic segmentation results compared with the 2D methods such as FCN and CRF-RNN. Figure \ref{compare} shows some semantic segmentation samples. Taking the advantages of the 3D constraints and refinement in our proposed framework, our semantic segmentation results are more uniform, sharp and accurate.

{\setlength{\tabcolsep}{1.5em}
\renewcommand{\arraystretch}{1.2}

\begin{table}[]
\centering
\caption{Accuracy study results compared with other 2D semantic segmentation algorithms}
\label{resultt}
\begin{tabular}{@{}lllll@{}}
\toprule
                     & \begin{tabular}[c]{@{}l@{}}pixel\\ acc.\end{tabular} & \begin{tabular}[c]{@{}l@{}}mean\\ acc.\end{tabular} & \begin{tabular}[c]{@{}l@{}}mean\\ IU\end{tabular} & \begin{tabular}[c]{@{}l@{}}f.w.\\ IU\end{tabular} \\ \midrule
FCN\cite{Shelhamer2017}                  & 81.61                                                & 63.69                                               & 49.54                                             & 76.16                                             \\
CRFRNN\cite{crfasrnn_iccv2015}               & 85.68                                                & 51.73                                               & 41.32                                             & 79.76                                             \\
\textbf{Ours\_noCRF} & \textbf{87.86}                                       & \textbf{70.69}                                      & \textbf{54.81}                                    & \textbf{81.86}                                    \\
\textbf{Ours}        & \textbf{89.42}                                       & \textbf{72.06}                                      & \textbf{56.32}                                    & \textbf{84.30}                                    \\ \bottomrule
\end{tabular}
\end{table}}

\subsection{User Experience Evaluation}

We conduct a user study to evaluate the effectiveness of the semantic-based MR system. Using the throwing plate game, three test conditions are designed by setting different collision responses: 

1) \textit{No Collision Mesh}: Virtual plates were thrown into the real world without any collision being detected. 

2) \textit{Uniform Collision Mesh}: Virtual plates interact with the real world with the uniform collision mesh being activated, but no object-specific interaction is generated. 

3) \textit{Semantic Collision Mesh}: Physics responses of the virtual plates with the real-world objects are dependent on the material properties of the objects in the real world. 

The objective of the user study is to assess the realism of the MR environment by measuring how much the semantic-based interactions matches the user anticipations. We investigate whether or not the semantic-based interactions can significantly improve the realism of MR systems and delivers immersive user experience. 

Firstly, we evaluate the realism of the physical interactions such as collision responses in MR systems. We test to see if users are able to detect differences in these three interaction conditions between virtual and real objects in short video clips, and whether or not the realism in MR can be improved via context-aware physical responses. Secondly, to ensure the quality of qualitative study, we test if there is any risk that the user experience of the proposed MR system could be affected by his or her previous engagement with computer games and MR or VR technologies. 


\subsubsection{Participants}

A questionnaire was designed and an online servery platform was used to conduct the user study. Anonymous participants were recruited without restrictions on age and gender. Each participant was firstly asked whether he/she had any previous experience with FPS games and VR/AR games, and then asked to watch the three video clips carefully. Each video clip can be viewed repeatedly, so that the participant can take time to digest and answer the questions. Each of the video clip was rated by participants on the scale of 1 (very bad) to 10 (very good) based on the quality of the MR interactions and realism.       



A total of 68 questionnaires were received, in which 6 responses were removed for reasons either participants did not confirm that they have watched the videos carefully or their viewing time was too short (less than 10 seconds) indicating little interest from the participants. Among the 62 valid questionnaires, 69.57\% had experience with FPS games, 65.22\% had experience with VR/AR games.

\subsubsection{Results}

\begin{figure}[]
\centering
\includegraphics[width=0.47\textwidth]{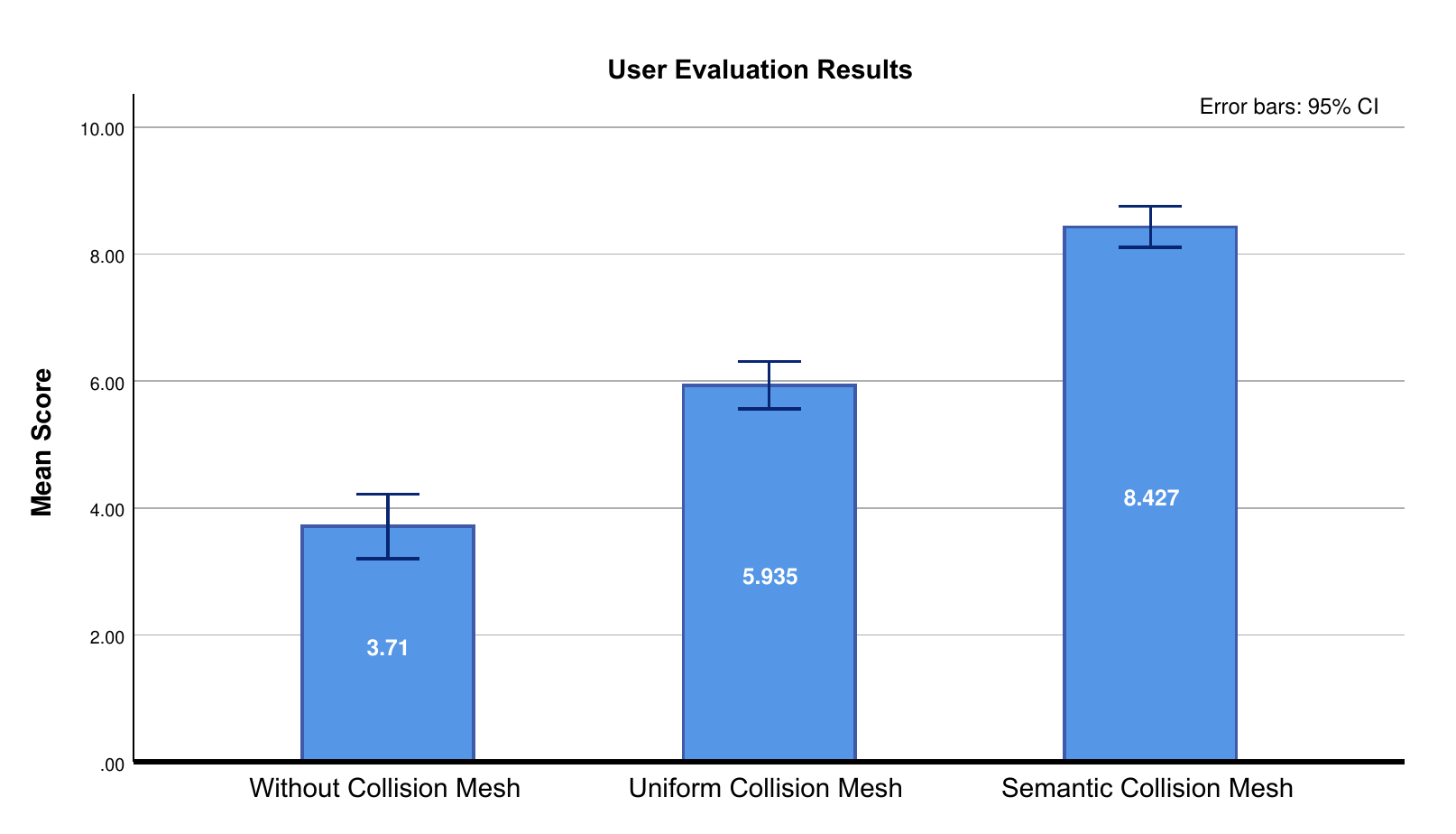}
\caption{The user experience evaluation results.}
\label{eva}       
\end{figure}

We used the score from 1 to 10 as the interval data so that we can use parametric ANOVA to analysis the data. We have performed a repeated measure ANOVA test to analyze scores obtained for the three conditions. Mauchly's test indicated that the assumption of sphericity had been violated ($X^{2}=42.029, p <0.001, \varepsilon=0.665$ ). Therefore, we used Greenhouse-Geisser to correct the degrees of freedom. Main effects were found within the three conditions ($F_{1.330,81.135}=212.293, p<0.001$). 

The following post hoc Bonferroni pairwise comparisons show that the \textit{Semantic Collision Mesh} ($M=8.427, SD=1.995$) is significantly better than the other two MR conditions ($p<0.001$), indicating that the proposed semantic interactions through the inference of material properties can greatly improve the realism of the MR system. We also found that the \textit{Uniform Collision Mesh} ($M=5.935, SD=1.458$) offers much better MR experience ($p<0.001$) than the \textit{No Collision Mesh} ($M=3.71, SD=1.99$) but less realistic compared with the semantic context-aware MR. The mean scores of the three system conditions are shown in Figure \ref{eva}. 

Furthermore, as a large number of our participants have either experienced FPS games (69.57\%) or VR/AR games (65.22\%), we also conducted a between-subjects repeated measure ANOVA test to reveal whether this experience has an influence on the user when assessing the results due to prior exposure to VR, MR, and games technologies. It has been shown that the final test results are not affected by either the experience of FPS games ($p=0.793$) or VR/AR games ($p=0.766$).

\subsection{User Feedback}
Many participants were interested in the MR system and gave very positive comments and feedback about their MR experience that the system provides. Comments are such as \textit{"This game (throwing plates) is amazing! I never experienced such MR experience before."}; some people commented on the importance of material-specific interactions even the low quality models, textures and animation being used in the current prototype, \textit{"Although the interaction sometimes is not very obvious, it really makes a lot of difference."}; some people criticized the MR system without semantic interaction: \textit{"The next second when the plates break when hitting a soft chair (indicating the MR system with Uniform Collision Mesh), I won't play it again, as it violates the basic physical law."}, while other people cannot wait to play our semantic interaction MR game \textit{"Your game creates a realistic interactive experience, nice work! When will you release your game?"}.


\section{Conclusion and Discussion}
We show how deep semantic scene understanding methodology combined with dense 3D scene reconstruction can build high-level context-aware highly interactive MR environment. Recognizing this, we implement a material-aware physical interactive MR environment to effectively demonstrate natural and realistic interactions between the real and the virtual objects. Our work is the first step towards the high-level interaction design in MR. This approach can lead to better system design and evaluation methodologies in this increasingly important technology field.

There are some immediate directions for future research and we mention two such directions now. Although in this paper we focus our discussions on material understanding and its semantic fusion with virtual scene in MR environment, the concept and the framework presented here are applicable to address many other context-aware interactions in MR, AR and even VR. The framework can be extended by replacing the training dataset with more general object detection databases for constructing different interaction mechanisms and context. Realistic physically-based sound propagation and physicall-based rendering using the proposed context-aware framework for MR are promising directions to pursue. Integrated with multi-modal interactions, immersive experience can be achieved. Our results have hinted that the study of semantic constructions in MR as a high-level interaction design tool is worth pursuing, as more comprehensive methodologies emerging, complex rich MR applications will be developed in the near future.    

We believe that AI is not only for autonomous tasks of machines and robots, but also is for the improvement of human decision making when interacting with the real world through virtual interactions.

\ifCLASSOPTIONcaptionsoff
  \newpage
\fi



\bibliographystyle{IEEEtran}
\bibliography{template}
%

%








\end{document}